\documentclass[conference]{IEEEtran}
\IEEEoverridecommandlockouts

\usepackage{cite}
\usepackage{hyperref}
\usepackage{amsmath,amssymb,amsfonts}
\usepackage{graphicx}
\usepackage{textcomp}
\usepackage{xcolor}
\def\BibTeX{{\rm B\kern-.05em{\sc i\kern-.025em b}\kern-.08em
    T\kern-.1667em\lower.7ex\hbox{E}\kern-.125emX}}

\usepackage{booktabs}
\usepackage[edges]{forest}
\forestset{
  direction switch/.style={
    for tree={edge+=thick, font=\sffamily},
    where level>=1{folder, grow'=0}{for children=forked edge},
    where level=3{}{draw},
  },
}
\usepackage{multirow}
\usepackage{amsmath}

\usepackage{stfloats}
\usepackage[linesnumbered,ruled,vlined]{algorithm2e}

\usepackage{arydshln}

\usepackage[font=footnotesize]{caption}
\usepackage{subcaption}
\DeclareCaptionFormat{subfig}{\figurename~#1#2#3}
\DeclareCaptionSubType*{figure}
\usepackage[T1]{fontenc}
\usepackage[utf8]{inputenc}
\usepackage{amsmath,amssymb}
\usepackage[russian,english]{babel}

\usepackage{draftwatermark}
\SetWatermarkLightness{ 0.85 }
\SetWatermarkText{Preprint}
\SetWatermarkScale{ 1 }

\begin{document}
\title{FedSym: Unleashing the Power of Entropy for Benchmarking the Algorithms for Federated Learning\\
\thanks{Identify applicable funding agency here. If none, delete this.}
}

\author{\IEEEauthorblockN{1\textsuperscript{st} S. Ensiye Kiyamousavi}
    \IEEEauthorblockA{\textit{GATE Institute} \\
    Sofia, Bulgaria \\
    ensiye.kiyamousavi@gate-ai.eu}
    \and
    \IEEEauthorblockN{2\textsuperscript{nd} Boris Kraychev}
    \IEEEauthorblockA{\textit{GATE Institute} \\
    Sofia, Bulgaria \\
    boris.kraychev@gate-ai.eu}
    \and
    \IEEEauthorblockN{3\textsuperscript{rd} Ivan Koychev}
    \IEEEauthorblockA{\textit{Sofia University ”St. Kliment Ohridski”} \\
    Sofia, Bulgaria \\
    koychev@fmi.uni-sofia.bg}
    }

\maketitle

\thispagestyle{plain}
\pagestyle{plain}

\begin{abstract}
Federated learning (FL) is a decentralized machine learning approach where independent learners process data privately. Its goal is to create a robust and accurate model by aggregating and retraining local models over multiple rounds. However, FL faces challenges regarding data heterogeneity and model aggregation effectiveness. In order to simulate real-world data, researchers use methods for data partitioning that transform a dataset designated for centralized learning into a group of sub-datasets suitable for distributed machine learning with different data heterogeneity. In this paper, we study the currently popular data partitioning techniques and visualize their main disadvantages: the lack of precision in the data diversity, which leads to unreliable heterogeneity indexes, and the inability to incrementally challenge the FL algorithms. To resolve this problem, we propose a method that leverages entropy and symmetry to construct 'the most challenging' and controllable data distributions with gradual difficulty. We introduce a metric to measure data heterogeneity among the learning agents and a transformation technique that divides any dataset into splits with precise data diversity. Through a comparative study, we demonstrate the superiority of our method over existing FL data partitioning approaches, showcasing its potential to challenge model aggregation algorithms. Experimental results indicate that our approach gradually challenges the FL strategies, and the models trained on FedSym distributions are more distinct.
\end{abstract}


\section{Introduction}
\par While centralized machine learning methods require large datasets to perform optimally, collecting data is time-consuming and costly. It necessitates large storage and computational capacity. Besides, privacy concerns and regulations such as GDPR \cite{voigt2017eu} have made it challenging to centrally aggregate data from various devices to train effective models.

\par
\begin{figure}
    \centering
    \includegraphics[width=0.33\textwidth]{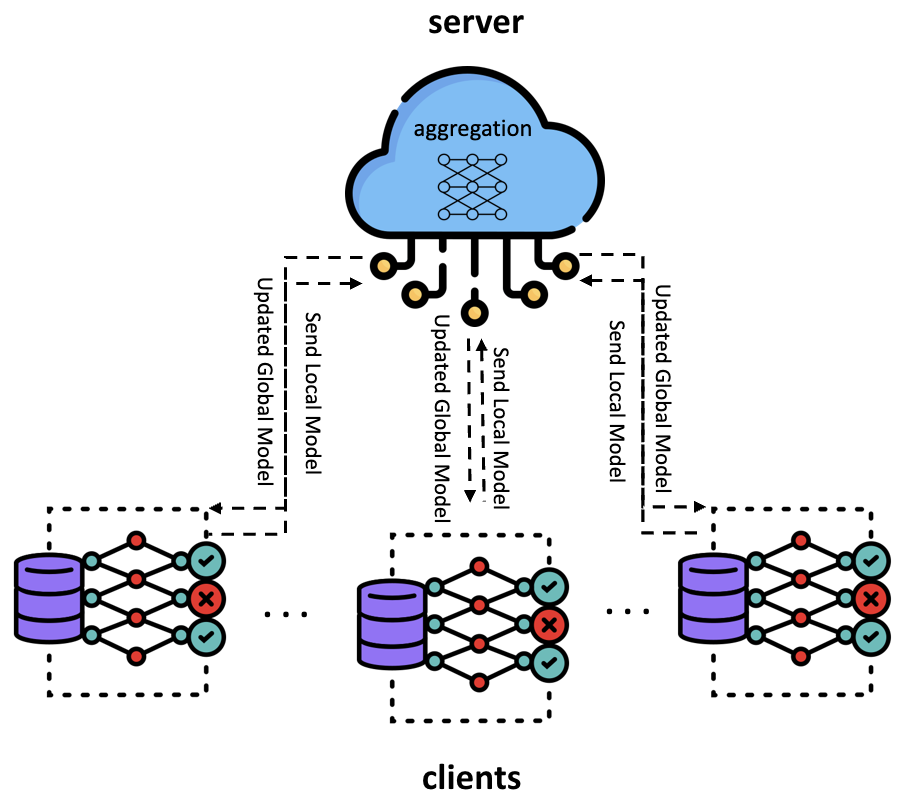}
    \caption{\smaller Schema of the federated learning process.}
    \label{fed}
\end{figure}

 To avoid direct access to datasets, FedAVG \cite{DBLP:journals/corr/McMahanMRA16} emerged as a fundamental algorithm for \emph{Federated Learning} (FL), proposed by \emph{Google} in 2016. This technology allows multiple agents to build a machine-learning model cooperatively without sharing local data. Figure \ref{fed} represents the standard schematic of a Federated learning system.

\par In recent years, many algorithms such as FedPROX \cite{li2018federated}, SCAFFOLD \cite{karimireddy2020scaffold}, Robust FL \cite{pillutla2019robust}, FEDOPT\cite{reddi2020adaptive}, Multi-center FL \cite{xie2020multi}, FedMA \cite{wang2020fedma}, FedDF \cite{lin2020ensemble} and others emerged, offering various advantages in different FL environments. 

\par All of these FL algorithms have promising strategies to train a shared model. However, in addition to the problem of locating an optimal aggregated model, other weaknesses in FL systems also need to be addressed.

\par In FL systems, common challenges include privacy\cite{mothukuri2021survey}, communication cost\cite{xu2021federated, konevcny2016federated}, and statistical heterogeneity\cite{karimireddy2020scaffold, huang2021personalized, zhu2021federated}. In this paper, we focus on statistical heterogeneity, which is crucial for the quality of the machine learning process.

\par Federated learning in a real-world setup involves multiple clients with non-independent and identical (non-IID) datasets, meaning that label distributions differ across learning agents. This non-IID constraint often significantly reduces federated learning systems' performance. While recent research in this domain has recognized this challenge, few studies have directly addressed the non-IID issue (e.g., \cite{huang2021personalized,karimireddy2020scaffold,DBLP:journals/corr/abs-2102-02079, no_fear,wang2022unreasonable, zhu2021federated,vahidian2022rethinking}). However, not all of these studies mentioned the importance of data partitioning settings to challenge FL algorithms.

Recently, \cite{DBLP:journals/corr/abs-2102-02079} did constructive studies and experiments using six non-IID data partitioning settings (The introduced categories are represented in chart \ref{chart1} and more details are provided in section \ref{background}). They also ranked \emph{state-of-the-art} FL algorithms on their categorized non-IID settings. Although their suggested partitioning strategies are very beneficial for the federated learning community, there is still a need to perform more experiments on data partitions generated by other types of strategies to help the FL algorithms be more applicable to a real-world scenario.

\par From the critical analysis of the experiments, we found a lack of attention to three factors: 
\begin{itemize}
\vspace{-1mm}
    \item First, we noticed that we need a metric to measure the amount of heterogeneity in each client’s local data. Such a metric can help better understand the non-IIDness in the clients' data. Although most of the recent papers use non-IIDness or imbalance measurement metric ($\alpha$ degree in \emph{Dirichlet distribution}), in this paper, we propose the usage of \emph{entropy balance} ($\beta$ degree) or Shannon's Evenness Measure, which is a normalized version of the well-known \emph{Shannon's Entropy} used by  \cite{vajapeyam2014understanding}, and \cite{magurran2003measuring} in measuring biological diversity.

    \item Second, we found that the introduced metric of entropy balance is in direct correlation with the difficulty of a training dataset, e.g., datasets with higher values of entropy balance produce models with higher accuracy. Moreover, to our knowledge, there are no methods for data partitioning that focus on the resulting entropy of the training datasets.

    \item Third, the existing partitioning strategies for generating non-IID datasets, also mentioned by \cite{DBLP:journals/corr/abs-2102-02079, zhu2021federated}, are classified as skewed distribution partitioning. Therefore, there is a lack of symmetrical distribution partitioning or, to our knowledge, any study to explore such a strategy.
\end{itemize}

\par \textbf{Contributions.} To the best of our knowledge, the described method in this paper is the first to guarantee: 
(i) equal entropy balance for all clients' training datasets
(ii) symmetrical class distributions for the clients and 
(iii) generated data partitions in a wide range of heterogeneity levels.
The goal of the method is to provide data partitions for precise benchmarking of the FL algorithms in various heterogeneity levels. This is achieved using entropy degree as a desired heterogeneity to generate client data partitions.

\par The remainder of this paper is structured as follows:
we provide the Background and Related Work in section \ref{background}. We introduce our proposed method, FedSym, in Section \ref{fedsdp}. Section \ref{experiments} presents the Experimental results. Section \ref{limitations} discusses the limitations of the method and the opportunities for future work. 

\section{Background and Related Work}
\label{background}

\par A typical assumption in machine learning is that observations in data sets are independently and identically distributed. \textbf{Independent and Identically Distributed (or IID) data} refers to a statistical property of a dataset where each data point is independent of the others and has the same distribution \cite{golden2020statistical} i.e., the data is generated from the same distribution and has no relationship between the data points.

\par Unfortunately, this assumption cannot be generalized to distributed machine learning, particularly federated learning. In addition, since FedAVG emerged, demonstrating the robustness of federated learning methods on heterogeneous datasets (unbalanced and non-IID data distributions) has been essential.

\par \textbf{Data heterogeneity} refers to the differences or inconsistencies in the data distribution, features, and labels among different datasets or sources. This can be seen in data structures, types, distributions, quality, and content differences. In our case, we use data heterogeneity to refer to the inconsistency of the distribution of classes in the training datasets.

Despite the claim made by the authors in \cite{McMahan2016} that FedAvg can handle data heterogeneity, a wealth of studies has shown that FL accuracy deteriorates when dealing with heterogeneous data \cite{zhao2018federated}. The primary cause of the performance decline is weight divergence in the local models. In other words, because of the diversity in local data distributions, local models with the same initial parameters will converge to different models. As a result, the FL process slows the convergence and worsens learning performance as the divergence between the averaged global model and the optimal model keeps rising. Let's assume we could aggregate the datasets of all clients joined in an FL system. Then, an optimal model is a model developed from a centralized learning process on an aggregated dataset.

To study more on this phenomenon, understand it better, and simulate a more real FL system, we need to use various non-IID \textbf{data partitioning strategies} for our experiments.

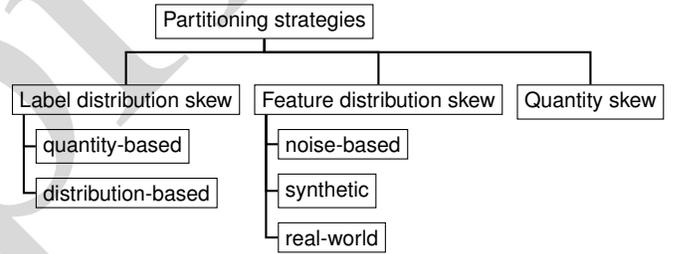
\begin{figure}[h!]
\centering
\smaller
\begin{forest}
  direction switch
  [Partitioning strategies
      [Label distribution skew
        [quantity-based]
        [distribution-based]]
      [Feature distribution skew
        [noise-based]
        [synthetic]
        [real-world]]
      [Quantity skew]
    ]
\end{forest}
\caption{Classification of the existing data partitioning strategies}
\label{chart1}
\end{figure}

\par Figure \ref{chart1} is a comprehensive summary of strategies for non-IID data partitioning defined by \cite{DBLP:journals/corr/abs-2102-02079}. Real federated learning datasets likely are a mixture of these scenarios and can not be directly categorized as one of these defined subgroups \cite{kairouz2019advances}. Recent research on federated learning tends to focus on label distribution skew \cite{li2020lotteryfl,qu2022rethinking, zhang2022federated} to simulate a non-IID system of datasets. They apply a partitioning strategy on an existing large dataset (same as the well-known CIFAR10\footnote{https://www.cs.toronto.edu/~kriz/cifar.html}) to provide a number of training datasets for the learning agents.

\par In \emph{Label distribution skew}, the class distributions vary across learning agents. It can be divided into two main strategies:

\begin{itemize}
    \item \emph{\textbf{Quantity-based label imbalance:}} In this setting, each party has a fixed number of data samples with a certain number of labels. This strategy was first introduced in FedAvg \cite{McMahan2016} and has been used in other studies like \cite{geyer2017differentially,li2018federated}. The authors propose a general partitioning strategy where each party is assigned a fixed number of different labels, and the samples for each label are randomly and equally divided among the parties.
    
    \item \emph{\textbf{Distribution-based label imbalance:}} In this setting, introduced by \cite{yurochkin2019bayesian} and based on the Dirichlet distribution, each client receives a proportion of training samples from each class, present in the original dataset. Specifically, they sample $p_{k} \sim DirN (\alpha)$ and distribute $p_{k,j}$ instances of class k to client j. Dir(·) stands for the Dirichlet distribution, and $\alpha$ is a concentration parameter $(\alpha > 0)$. 
\end{itemize}

\par
\begin{figure*}[!ht]
  \begin{subfigure}[b]{.5\linewidth}
    \centering
    \includegraphics[height=5.5cm, width=6.0cm]{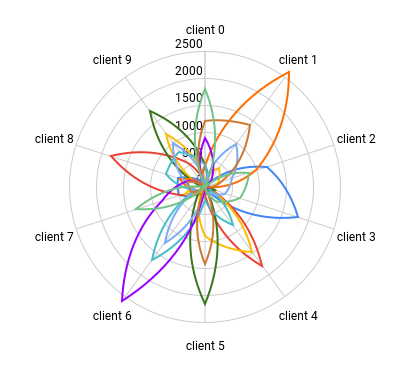}
    \caption{Labels distribution per client for Dirichlet dist. $\alpha = 0.7$}
    \label{fig:rad_a05}
  \end{subfigure}%
  \hspace*{5mm}
  \begin{subfigure}[b]{.5\linewidth}
    \centering
    \includegraphics[height=5.5cm, width=6.5cm]{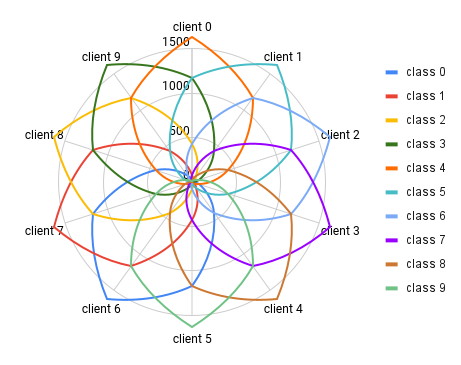}
    \subcaption{Labels distribution per client for FedSym dist. $\beta = 0.7$.}
    \label{fig:rad_b07}
  \end{subfigure}
\end{figure*}

 Many recent studies like \cite{karimireddy2020scaffold,wang2020federated, lin2020ensemble, hsu2019measuring, li2020practical, wang2020tackling, acar2021federated,awan2021contra,fraboni2021clustered, gao2022feddc, liu2022energy,sturluson2021fedrad} have adopted \emph{Distribution-based label imbalance}(Dirichlet partitioning strategy) since it was first used in \cite{yurochkin2019bayesian}. Figure \ref{fig:rad_a05} shows an example of this partitioning strategy.
Therefore, it will also be our center of attention in this paper. We will compare the accuracy results of baseline FL algorithms on data generated by the well-known Dirichlet method and the proposed FedSym.

Our study uses widely the following metric that we would like to remind:
\par \textbf{Entropy Balance} or \textbf{Shannon's Evenness Measure} is a metric to measure the amount of diversity or imbalance for each client’s local data. We can use \emph{Shannon’s Entropy}\cite{magurran2003measuring} for a given class distribution $C$:

    \begin{quote}
        ``\textbf{Shannon’s entropy} is an amount of the uncertainty related to a random variable. It expresses the quantity of the information contained in a message.''
    \end{quote}
    \begin{equation}\label{shannonInd}
    H(C) = -\sum_{c\in C}p(c)\log_{2}{p(c)}
    \end{equation}

The maximum diversity ($H_{max}$) would be reached when all labels have equal abundances. In other words, when $H = H_{max} = \log_{2}{|C|}$.\cite{magurran2003measuring}
Based on this definition, The ratio of observed
diversity to maximum diversity can be used to measure evenness or heterogeneity \cite{magurran2003measuring}. Therefore, we can define \emph{Shannon's Evenness Measure} as
\begin{equation}\label{eqInd}
\beta(C) = \frac{-\sum_{c\in C}p(c)\log_{2}{p(c)}}{\log_2{|C|}}
\end{equation}

that is a normalized form of Shannon's Entropy. In this paper, for simplification, we call it \textbf{\emph{entropy balance ($\beta$)}}. 

The values of $\beta$ vary from 0, denoting a completely imbalanced dataset, to 1 for a completely balanced one, e.g., $\beta \in [0,1]$. To generate datasets with equal entropy balance, we start from an utterly balanced dataset and apply Gaussian distribution to limit the presence of certain classes and decrease the entropy of the data samples.

\section{Symmetrical data partitioning using Shannon's Entropy}
\label{fedsdp}

\par \textbf{Motivation.} To effectively train a model in machine learning, the training dataset must have a fair balance of class samples. However, in real-world scenarios of federated learning, the class distribution is often imbalanced across the different models for various reasons. For example, in a hospital setting, there may be more cases of a certain medical condition in one hospital than in another. As a result, this can impact the overall accuracy of the FL system. For this reason, we need to provide a non-IID dataset for each client. Since we don’t have access to actual federated learning datasets, we should choose a data partitioning strategy that, from a given dataset, generates non-IID sub-datasets with a varying range of heterogeneity. In this way, we can determine if an FL method is suitable to deal with class-imbalanced datasets. Most recent papers in federated learning use the Dirichlet distribution data partitioning strategy to generate non-IID datasets.  The method controls the imbalance using $\alpha > 0$. However, the generated datasets with different $\alpha$ degrees have overlapping levels of data diversity. Having a metric to assess the imbalance and diversity of the data and the ability to generate precisely imbalanced partitions is critical for FL algorithm evaluation. 
Entropy and its normalized version, entropy balance ($\beta \in [0, 1]$), measure the amount of information contained in a dataset. This metric can be used to evaluate the variety of the data within a dataset and indicate the difficulty degree of a data partition for FL strategy benchmarking. To implement this idea, we propose \textbf{FedSym}, an entropy-based data partitioning strategy that provides datasets with strict $\beta$ degrees. Our intuition is that by varying the entropy metric of the datasets, we can achieve gradually challenging collections of datasets and, ultimately, obtain the 'most difficult' data partitions.  For a target entropy balance $\beta$, the method computes a variance as an input to a discrete Gaussian distribution PMF that is then applied to deliver the exact count of elements per class required in each client's dataset.
After finding the optimal Gaussian distribution as an array of per-class numbers, we can apply rotation to generate the desired distributions for all learning agents (clients). The result is symmetrically identical data distributions for each client, with equal entropy balance (cf. Fig. \ref{fig:rad_b07}).

\par \textbf{Entropy balance and the standard deviation of a Gaussian distribution.} The goal of data partitioning for an FL training process is to start from a common dataset suitable for centralized machine learning and split it to $k$ clients or learners. 
We plan to use a Gaussian distribution with precisely calculated standard deviation to propose a per-class number of samples with a desired entropy balance value. Therefore, we can define the following \textbf{problem setup}:

For a dataset $D$ containing samples from $l$ classes (labels), we have to provide a set of $k$ class distributions - one for each learning agent (client), where each class distribution is in the form  $C_i=[p(c_0), p(c_1), ..., p(c_{l-1})], i \in [0, k-1]$. Each $C_i$ contains $l$ per-class numbers of samples, estimated by a discrete Gaussian Distribution with varying mean value $\mu$ and identical standard deviation $\sigma$. For simplicity, we can also assume that $c_i = i$ e.g. we are searching for a standard deviation $\sigma$ that is used in a discrete Gaussian distribution over the numbers in the range $0,...,l-1$, and the values of its PMF generate a training dataset with a desired entropy balance. Once the desired standard deviation is found, we can use the mean value $\mu$ of the Gaussian distribution to replicate it symmetrically over $k$ clients.

To solve the above problem, we study the relationship between the \emph{entropy balance} and the variance of a \emph{Gaussian Distribution}.
Since Gaussian Distributions are continuous by nature and label distributions are discrete and finite, we use its discrete implementation as defined by \cite{canonne2020discrete, andrich2021rasch}:
\begin{quote}
\textbf{Discrete Gaussian Distribution:} Consider parameters $\mu, \sigma \in \mathbb{R}$ such that $\sigma>0$. The \emph{Discrete Gaussian Distribution} with mean $\mu$ and variance $\nu =\sigma^{2}$ is represented as $\mathcal{N}_{\mathbb{Z}}(\mu, \sigma^{2})$. This distribution is defined over integers and can be expressed as:

\begin{equation}
\label{disceret_Gauss}
\centering
\forall x \in \mathbb{Z}, \underset {x \leftarrow \mathcal{N}_{\mathbb{Z}}(\mu, \sigma^{2})} {\mathbb{P}} [X=x] = \frac{1}{\gamma} e^{- \frac{1}{2} (\frac{x-\mu}{\sigma})^{2}}, 
\
\ \ \ \ \gamma = \sum_{y \in \mathbb{Z}} e^{- \frac{1}{2} (\frac{y-\mu}{\sigma})^{2}} \ \ \ \
\end{equation}
\\
\end{quote}

\par The key distinction between the discrete and continuous Gaussian distributions lies in the normalization constant $\gamma$. For the continuous Gaussian(CG) distribution, the constant is given by $\gamma = \sigma\sqrt{2\pi}$, ensuring that the integral of the density function over its entire range equals 1. In contrast, for the discrete Gaussian (DG), $\gamma$ ensures that the sum of the probabilities equals 1, and $\gamma_{CG}$ serves as an upper limit of $\gamma_{DG}$. $\gamma_{DG}$ is required to sum an infinite (in $\mathbb{Z}$) or finite series of probabilities to estimate it. As a result, we have $\gamma_{DG} < \gamma_{CG}=\sigma\sqrt{2\pi}$. 

\par From equation \ref{eqInd}, the \emph{entropy balance} of $C$ is calculated as follows: 
 \[\beta(C) = \frac{-\sum_{i=0}^{l-1}p(c_i)\log_{2}{p(c_i)}}{\log_2{|C|}}\]
Where $|C|=l$ is the number of classes available in the original dataset $D$ 
and $p(c_i)$ is the \emph{PMF} of a discrete Gaussian distribution as defined in eq.\ref{disceret_Gauss}. As a reminder: we assume $c_i=i \in [0, ..., l-1]$.
\begin{equation}
\label{semi_discret}
    p(i) = \frac{1}{\hat{\gamma}} e^{- \frac{1}{2} (\frac{i-\mu}{\sigma})^{2}}, \ \ \ \ \hat{\gamma} = \sum_{i=0}^{l-1}e^{- \frac{1}{2} (\frac{i-\mu}{\sigma})^{2}}<\sigma\sqrt{2\pi}
\end{equation}
Therefore, we can rework the expression for entropy balance as follows: 
\begin{equation}
 \begin{array}{l}
\beta (C) = \frac{-1}{\log_{2}{|C|}}[\sum_{i=0}^{l-1} p(i)\log_{2}(\frac{1}{\hat{\gamma} } e^{-\frac{1}{2} (\frac{i-\mu }{\sigma } )^{2}})] =\\
\\
= \frac{-1}{\log_{2}{|C|}}[\sum_{i=0}^{l-1} p(i)\log_{2}\frac{1}{\hat{\gamma} } - (\frac{1}{2\sigma^2}\log_2{e}).(\sum_{i=0}^{l-1}p(i).(i-\mu)^2)]
\end{array}
\label{eq4}
\end{equation}

\cite{stewart2009probability}\footnote{ On page 89} shows that the mean and variance of a discrete random variable can be computed as follows:
        \begin{align*}
            & \mu = \Sigma_{c \in C} p(c)c , \ \ \ \ \ \ \ \ \ \ \nu = \sigma^2 = \Sigma_{c \in C} (c-\mu)^{2}p(c)
        \end{align*}
Therefore, we can simplify \ref{eq4} and express the entropy balance of a class distribution $C$ as:
\begin{equation}
\begin{array}{l}
\beta (C) = \frac{-1}{\log_{2}{|C|}}[1.\log_{2}\frac{1}{\hat{\gamma} } - (\frac{1}{2\sigma^2}\log_2{e}).\sigma^2]=\\
\\
\ \ \ \ \ \ \ \ \ \ =\frac{1}{2.\log_{2} |C|} (\log_{2} \hat{\gamma^2}+\log_{2} e)=\frac{1}{2}\log_{|C|} (e\hat{\gamma^2})
\end{array}
\label{eq5}
\end{equation}

As $\hat{\gamma} < \sigma\sqrt{2\pi}$, we can conclude that $\beta(C) < \frac{1}{2}\log_{|C|} (e2\pi\sigma^2)$.
We can notice in equation \ref{eq5} that the mean $\mu$ does not influence the entropy balance, so as shown by \cite{conrad2004probability}, we can conclude that all Gaussian distributions with identical $\sigma$ have identical entropy over the same discrete input. Also, the entropy of a discrete Gaussian distribution is an increasing function of its variance, because $\hat{\gamma}$ is an increasing function of $\sigma$ for given input and mean value $\mu$ (\ref{semi_discret}).             

\par Furthermore, we can reverse equation \ref{eq5} to obtain a lower limit of the variance $\nu$ and the standard deviation $\sigma$ by the entropy balance $\beta$:
\begin{equation}
\label{eq_sigma}
\nu = \sigma^{2} > \frac{|C|^{2\beta(C)}}{2\pi e}
\end{equation}

which is a lower approximation for the standard deviation of the discrete Gaussian distribution that we are searching for.

\begin{algorithm}
\SetNoFillComment
\LinesNumbered
\smaller
\SetArgSty{textnormal}
\KwIn{$D$: a dataset used for training,\\ 
    $l$: number of class labels (derived from the dataset), \\
    $k$: number of clients,\\
    $\beta$: the goal for entropy balance,\\
    $\epsilon$: allowed error tolerance
}
\KwOut{$C$: a label distribution - a list of $l$ per-class number of samples representing a data partition for a single client.}
\SetKwFunction{FMain}{GaussianPartition}
\SetKwFunction{FSecond}{EntropyBalance}

\BlankLine

$\mu \gets k/2$

$S \gets \frac{|D|}{floor(k/l)+1}$  \textcolor{blue}{\emph{num of training samples per client}}

$\sigma_0 \gets \sqrt{\frac{|C|^{2\beta}}{2\pi e}}$ \ \ \textcolor{blue}{\emph{from eq. \ref{eq_sigma}}}


%
$\hat{\beta} \gets -1$

\While {$|\beta - \hat{\beta}|\geq\epsilon$}{

    
    $C \leftarrow$ \FMain{$\mu$,$\sigma_i$, $l$, $S$}

    $\hat{\beta} \gets \FSecond{C}$

    $t_i \gets \frac{1}{\ln{|C|}\times\sigma_i}$ \ \ \textcolor{blue}{\smaller \emph{= $\beta'(\sigma_i)$, first derivative of eq.\ref{eq5} at $\sigma_i$}}
    
    $\sigma_{i+1} \gets \sigma_i + \frac{(\beta - \hat{\beta})}{t_i}$ \ \ \textcolor{blue}{\smaller \emph{update $\sigma_i$ using the tangent $t_i$}}
}
\Return $C$

\BlankLine



              

\SetKwProg{Fn}{Function}{:}{}

\Fn{\FMain{$\mu$,$\sigma$,$l$,$S$}}{
\ \ \ \ \ \ $C \gets []$ 

\ \ \ \ \ \ \For{$i = 0$ \KwTo $l-1$}

\ \ \ \ \ \ \ \ \ \ $C[i] \gets p(i)*S$
}

\BlankLine
\caption{FedSym - finding the partition with the closest entropy balance}
\label{alg:fedsdp}

\end{algorithm}

\par \textbf{FedSym.} The method for achieving class distributions with equal entropy balance is shown in algorithm \ref{alg:fedsdp}. It takes as an input a training dataset $D$ and derives the number of classes $l$ from it. The dataset could be any popular one like CIFAR10, CINIC10 \footnote{https://datashare.ed.ac.uk/handle/10283/3192} or any other. Inputs also include the number of clients $k$, the targeted entropy balance $\beta$, and a small value for error tolerance $\epsilon$.

\begin{figure}[!ht]
    \begin{center}
        \includegraphics[width=0.32\textwidth]{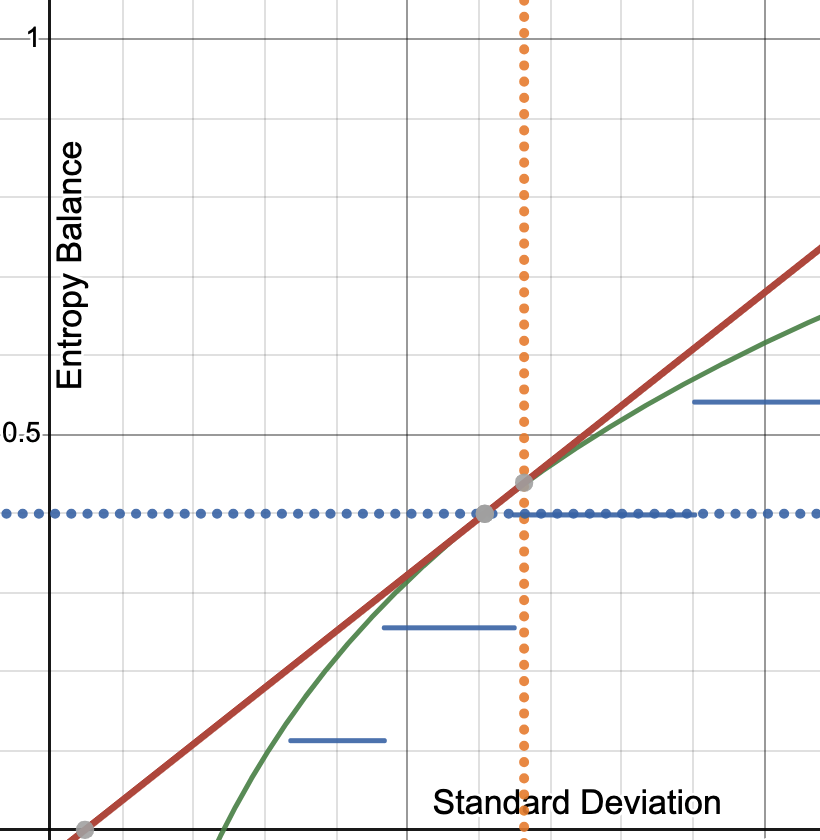}
    \end{center}
    \caption{Finding $\sigma$ by using continuous Gaussian distribution as guidance}
    \label{fig:FedSym}
\end{figure}

The algorithm aims to find the standard deviation $\sigma$ that produces a data partition with the closest possible entropy balance to the target $\beta$ by using the function of a continuous Gaussian distribution and its first derivative as guidance. A sample calculation is shown in figure \ref{fig:FedSym}, where we see the continuous distribution in solid green, its derivative at $\sigma_0$ in solid red, and the discrete Gaussian distribution in solid blue. The target entropy balance is present as a dotted blue line, while the solution $\sigma_1$ is a dotted orange line.

\begin{table*}[b]
\begin{center}
\fontsize{10}{8}\selectfont
\resizebox{16.5cm}{!}{%
\begin{tabular}{c c c c c c c c c c c c } 
 \specialrule{0.05em}{0.05em}{0.2em} 
 \multicolumn{1}{c}{}&
 \multicolumn{10}{c}{Range of the entropy balance for all clients' data}\\
 \cmidrule(r){2-11}
         & 0.1 & 0.2 & 0.3& 0.4& 0.5& 0.6 & 0.7 & 0.8 & 0.9 & 1.0 \\
\specialrule{0.05em}{0.05em}{0.5em}
Dirichlet ($\alpha$) & $0.07$--$0.53$ & $0.30$--$0.67$ & $0.36$--$0.60$ & $0.41$--$0.86$ & $0.49$--$0.87$ & $0.58$--$0.84$ & $0.59$--$0.87$ & $0.45$--$0.90$ & $0.52$--$0.88$ & $0.74$--$0.93$ \\
\specialrule{0.05em}{0.05em}{0.5em}
 FedSym ($\beta$) & $0.10$ &$0.20$ &$0.30$ &$0.40$ &$0.50$ &$0.60$ &$0.70$ &$0.80$ &$0.90$ &$1.00$ \\ 
\specialrule{0.05em}{0.05em}{0.5em}
\end{tabular}}
\end{center}
\caption{\smaller Range of the clients' entropy balance for data partitions generated by Dirichlet and FedSym methods \\
for 10 clients and on CIFAR10. $\alpha, \beta \in [0.1, 1.0]$.}
\label{tab:alpha_range}
\end{table*}

In the beginning, we initialize $\sigma$ using equation \ref{eq_sigma} and adjust it in a while loop, using the function's tangent for $\beta$ from equation \ref{eq5}.
The adjustment of $\sigma$ or $\Delta\sigma$ is estimated by dividing the subtraction of the achieved entropy balance $\hat{\beta}$ and the target $\beta$ by the value of the first derivative of the function from equation \ref{eq5}, calculated at the current point $\sigma$.

The result of the algorithm is a list of per-class number of samples representing a data partition for a single learning client. The entropy balance of this list is the closest possible number to the desired $\beta$. Next, to achieve symmetrical data partitioning, we rotate the generated list $k$ times in a loop and assign the obtained lists to each client. The final result is visualized in figure \ref{fig:rad_b07}, which illustrates the symmetry and the entropy equality, as opposed to the corresponding Dirichlet distribution in figure \ref{fig:rad_a05}. As a last step, using the generated lists, the algorithm selects unique samples from the original dataset to generate $k$ training sets by guaranteeing that each sample will be present only once. 

\section{Experiments and Results}
\label{experiments}

\par To examine the outcome of FedSym, we plan three experiments. The first one will compare the entropy of the produced data splits from FedSym with another method, namely the Dirichlet data partitioning. Next, we will examine the performance of common algorithms for federated learning over 20 dataset partitions to compare their performance. Finally, we will cross-compare the output of the resulting models via \emph{Centered Kernel Alignment (CKA)} to visualize the variety of the generated models by each heterogeneity index $\alpha$ and $\beta$.

\par \textbf{Entropy balance comparison.}
The first experiment, represented in table \ref{tab:alpha_range}, compares the entropy of the training datasets generated by FedSym and the popular Dirichlet method to estimate their final training difficulty.
\begin{table*}[!hb]
\begin{center}
\tiny
\resizebox{16.5cm}{!}{%
\begin{tabular}{l c c c c c c c c c c c } 
 \hline
 \multicolumn{1}{c}{}&
 \multicolumn{10}{c}{Heterogeneity index}\\
 \cmidrule(r){2-11}
         & 0.1 & 0.2 & 0.3& 0.4& 0.5& 0.6 & 0.7 & 0.8 & 0.9 & 1.0 \\
 \hline\hline
FedAVG on $\alpha$ &26.55\% &34.96\% &43.63\% &51.02\% &55.15\% &56.53\%&62.48\% & 61.41\% & 57.98\% &69.67\% \\ \hline
FedPROX on $\alpha$ &24.19\% &31.88\% &41.55\% &47.77\% &50.32\% &51.95\% &58.66\% &59.21\% &55.14\% &66.8\% \\ \hline
SCAFFOLD on $\alpha$ &32.56\% &43.08\% &46.06\% &60.03\% &60.02\% &61.54\% &67.65\% &66.94\% &68.24\% &70.79\% \\ \hline \hline
FedAVG on $\beta$ &11.04\% &17.74\% &31.40\% &45.01\% &51.99\% &62.85\% &66.89\% &69.15\% & 71.40\% & 71.78\% \\ \hline
FedPROX on $\beta$ &10.01\% &14.21\% &33.97\% &39.72\% &51.27\% &60.89\% &64.3\% &66.85\% & 68.93\% & 68.38\% \\ \hline
SCAFFOLD on $\beta$ &13.88\% &25.15\% &38.19\% &41.35\% &59.33\% &65.99\% &67.91\% &70.59\% & 70.58\% & 71.35\% \\ \hline\hline
\end{tabular}
}
\end{center}
\caption{Final accuracy for $\alpha$ and $\beta \in [0.1, 1.0]$ and FL algorithm.}
\label{tab:benchmark}
\end{table*}

 \begin{figure}[!ht]
    \vspace*{-0.2cm}
    \hspace*{-0.5cm}
    \centering
        \includegraphics[width=0.49\textwidth]{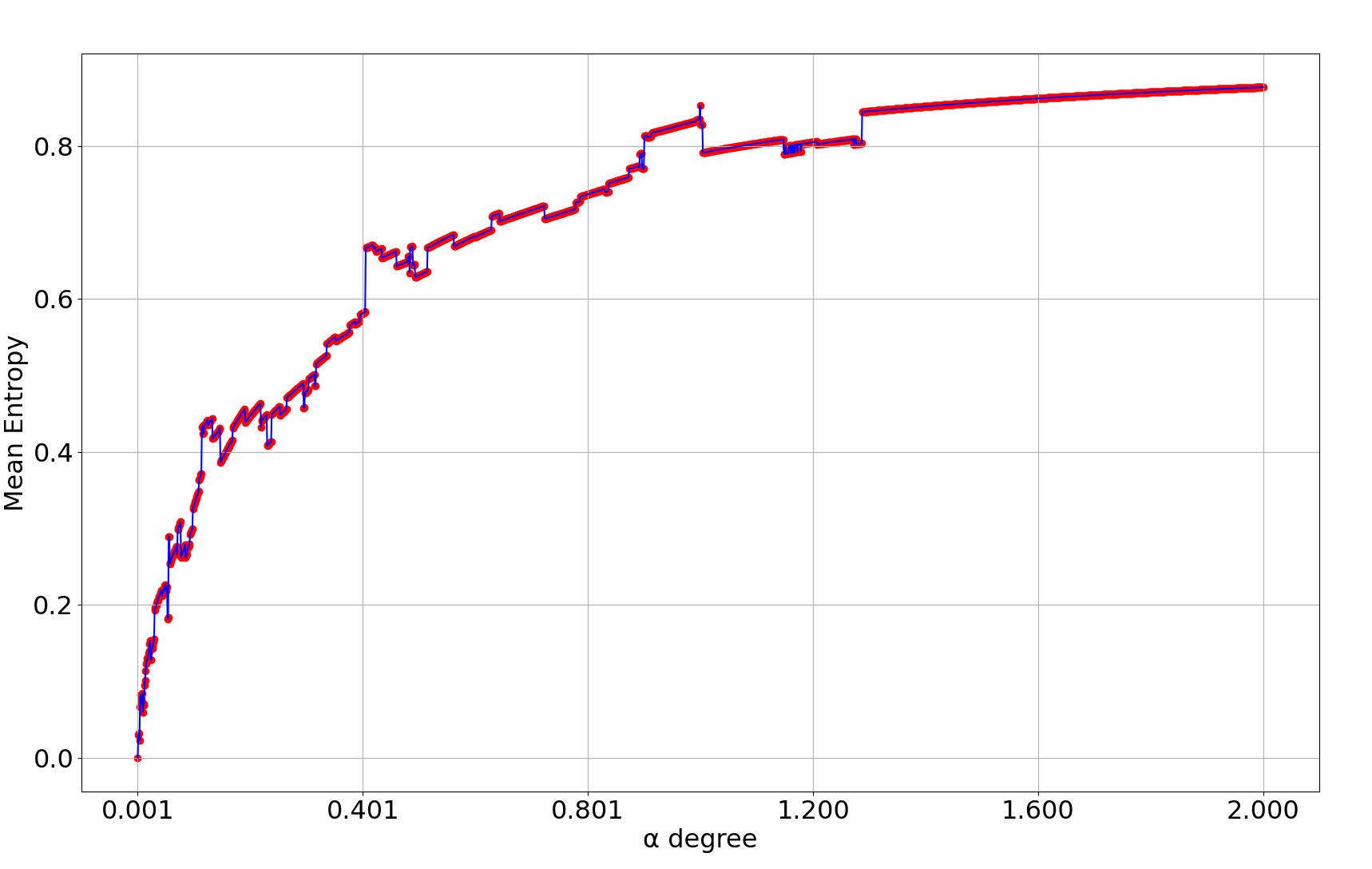}
    \caption{\smaller Mean Entropy balance of distributions generated using $\alpha \in [0,2]$}
    \label{fig:alpha_vs_beta_0-2}
\end{figure}
 We chose a popular image classification dataset with 50K training samples. The dataset name is not essential, but this could be the CIFAR10, and we distribute the data to 10 learning agents, a popular FL benchmarking configuration\cite{no_fear}. We measure the class diversity in the generated datasets using the \emph{entropy balance} metric. Then, for each index $\alpha$ (for the Dirichlet method), we find the minimum and maximum calculated entropy balances of all ten clients.
 It is important to remind that while the values of $\beta$ are limited to the interval $[0, 1]$, $\alpha$ can be any positive real number, and as shown by \cite{hsu2019measuring}, as its value advances towards infinity, the entropy balance of the produced data partitions is approaching $1.0$.
 
Table \ref{tab:alpha_range}, represents the range of the clients' entropy balance for data partitions generated by Dirichlet and FedSym methods for 10 clients on CIFAR10. 
We observe that for the Dirichlet distributions, the entropy balance values are spread over a wide range and these ranges (sometimes completely) overlap.
In general, this observation in Table \ref{tab:alpha_range} and \ref{table_eb}, suggests a direct relationship between the parameter $\alpha$ and the mean entropy balance. From \cite{hsu2019measuring}, we also know that with $\alpha \to \infty$, the distributions of all clients become more similar to the prior distribution. In contrast, with $\alpha \to 0$, each client holds examples from only one class. 
\begin{figure}[!ht]
    \vspace*{-0.2cm}
    \hspace*{-0.5cm}
    \centering
    \includegraphics[width=0.5\textwidth]{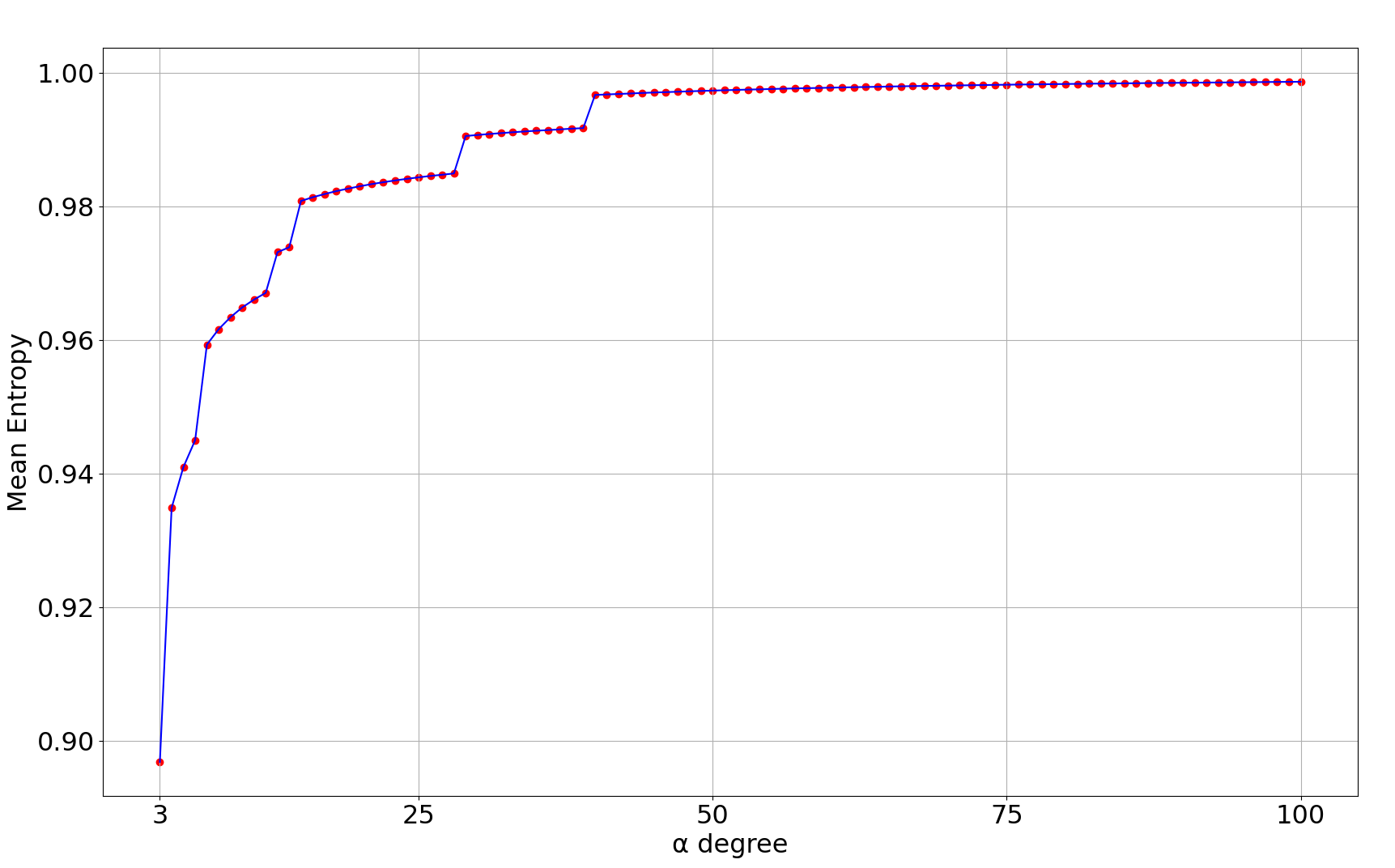}
    \caption{\smaller Mean Entropy balance of distributions generated using $\alpha\in [3,100]$}
    \label{fig:alpha_vs_beta_3-100}
\end{figure}
Because of the range overlaps and similarity of mean entropy for different $\alpha$ degrees, we investigated further alpha degrees. We chose $\alpha \in [0.001, 100]$, corresponding to the mean $\beta$ in the range $[0,1)$. We generated data distributions for 10 clients at each $\alpha$ increment: for values within $[0.001,2]$ at intervals of $0.001$ (i.e., $[0.001,0.002,...,2]$) and for those within $[3,100]$ (i.e., $[3,4,5,...,100]$). Then, we calculated the mean entropy for each alpha. The results of the computed mean entropy balances are graphically presented in figures \ref{fig:alpha_vs_beta_0-2} and \ref{fig:alpha_vs_beta_3-100}. At first sight, a clear pattern emerges: there is a discernible increase in the mean entropy as the parameter $\alpha$ grows. However, it's important to note that anomalies appear across various $\alpha$ values, mostly when $\alpha < 1$, creating interesting deviations in the observed patterns. As a result, these observations disprove the claim that there is a direct relationship between the parameter $\alpha$ and the data diversity, specially when $\alpha < 1 $. 



\begin{figure*}[!ht]
\hspace*{-2.2cm}
  \begin{subfigure}[b]{.4\linewidth}
    \centering
    \includegraphics[height=4.0cm, width=4.5cm]{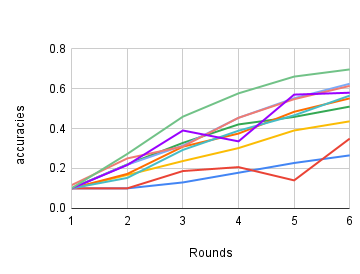}
    \subcaption{FedAVG on Dirichlet distributions}
    \label{fig:avg_dir_acc}
  \end{subfigure}%
  \begin{subfigure}[b]{.4\linewidth}
    \centering
    \includegraphics[height=4.0cm, width=4.5cm]{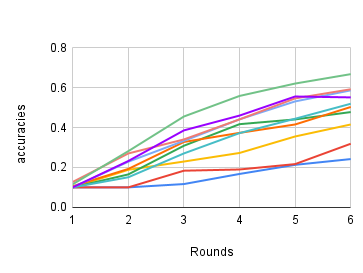}
    \subcaption{FedProx on Dirichlet distributions}
    \label{fig:prox_dir_acc}
  \end{subfigure}%
  \begin{subfigure}[b]{.4\linewidth}
    \centering
    \includegraphics[height=4.0cm, width=5.0cm]{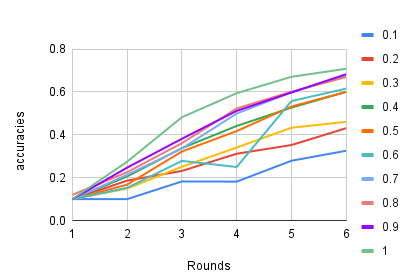}
    \subcaption{SCAFFOLD on Dirichlet distributions}
    \label{fig:scaffold_dir_acc}
  \end{subfigure}
\end{figure*}

\begin{figure*}[!ht]
\hspace*{-2.5cm}
  \begin{subfigure}[b]{.4\linewidth}
    \centering
    \includegraphics[height=4.0cm, width=4.5cm]{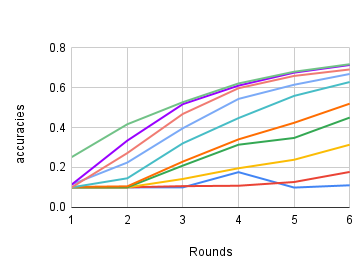}
    \subcaption{FedAVG on FedSym distributions}
    \label{fig:avg_ent_acc}
  \end{subfigure}%
  \hspace*{0.2cm}
  \begin{subfigure}[b]{.4\linewidth}
    \centering
    \includegraphics[height=4.0cm, width=4.5cm]{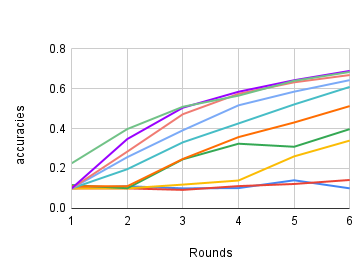}
    \subcaption{FedProx on FedSym distributions}
    \label{fig:prox_ent_acc}
  \end{subfigure}%
  \hspace*{0.2cm}
  \begin{subfigure}[b]{.4\linewidth}
    \centering
    \includegraphics[height=4.0cm, width=5.0cm]{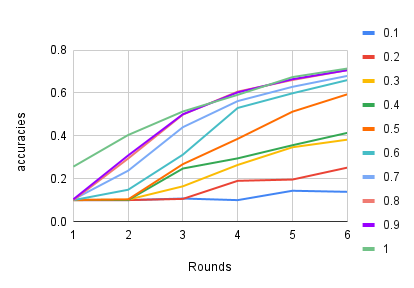}
    \subcaption{SCAFFOLD on FedSym distributions}
    \label{fig:scaffold_ent_acc}
  \end{subfigure}
\end{figure*}

\par \textbf{Impact of entropy balance on federated learning algorithms' performance.} In this experiment, we explore the relationship between the entropy balance of the training datasets and the performance of three algorithms for federated learning, namely, FedAVG, FedProx, and SCAFFOLD. We hypothesize that the value of entropy balance will determine the overall performance of the machine learning process. We run experiments over 20 data partitions generated by $\alpha \in [0.1, 1.0]$ and $\beta \in [0.1, 1.0]$.
We use a popular FL benchmarking configuration, based on the CIFAR10 dataset partitioned into ten subsets for ten learning clients, a convolutional neural network based on the model VGG11, and we run $3\times20$ complete FL training routines to verify the performance of all three FL algorithms over the 20 sets of data partitions. The data partitions are identical to the ones used in the previous experiment. 

In our experiments, we configured the learning rate ($lr$) at 0.016 and batch size to 50. We applied a decreasing rate of 0.95 to gradually reduce the $lr$ per round and employed a momentum of 0.9. For FedProx, the $\mu$ parameter was established at 0.01.

\par Table \ref{tab:benchmark} represents the obtained classification accuracy per FL algorithm and heterogeneity indexes $\alpha$ and $\beta$. Figures \ref{fig:avg_dir_acc}, \ref{fig:prox_dir_acc}, and \ref{fig:scaffold_dir_acc} present the performance of the 3 algorithms on all Dirichlet-based data distributions, while figures \ref{fig:avg_ent_acc}, \ref{fig:prox_ent_acc}, and \ref{fig:scaffold_ent_acc} contain the per-round accuracies of the aggregated models for the FedSym partitioning method. The results indicate a clear relationship between the value of the mean entropy balance and the overall training process and outcomes. We observe unstable and similar performance for the training on datasets with $\alpha \in [0.4, 0.9]$. At the same time, the data partitions generated by FedSym determine the results of the federated learning process. The final model accuracies are evenly spread in the range of 10\% to 70\% (figures \ref{fig:avg_ent_acc}, \ref{fig:prox_ent_acc}, and \ref{fig:scaffold_ent_acc} ).

\begin{figure*}[!ht]
\hspace*{1cm}
  \begin{subfigure}[b]{.3\linewidth}
    \centering
    \includegraphics[height=3.0cm, width=4.0cm]{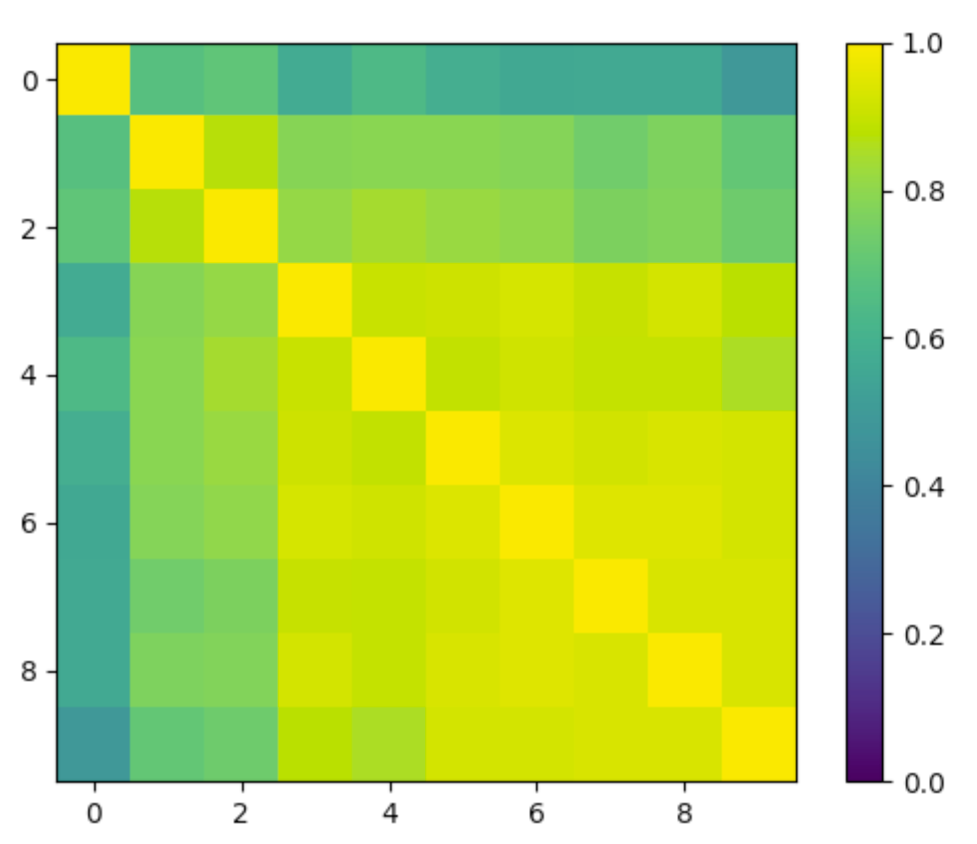}
    \subcaption{\small FedAvg $\alpha$}
    \label{fig:cka_dirichlet_fedavg}
  \end{subfigure}%
  \begin{subfigure}[b]{.3\linewidth}
    \centering
    \includegraphics[height=3.0cm, width=4.0cm]{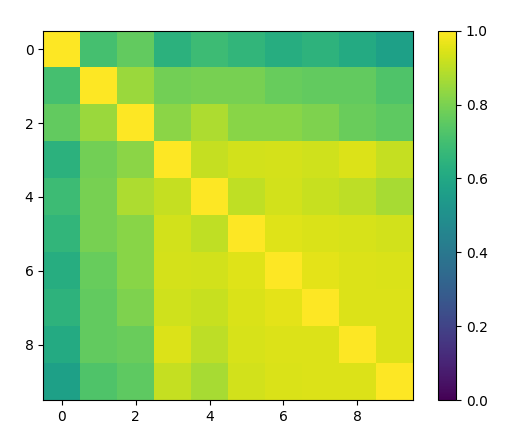}
    \subcaption{\small FedProx $\alpha$}
    \label{fig:cka_dirichlet_fedprox}
  \end{subfigure}%
  \begin{subfigure}[b]{.3\linewidth}
    \centering
    \includegraphics[height=3.0cm, width=4.0cm]{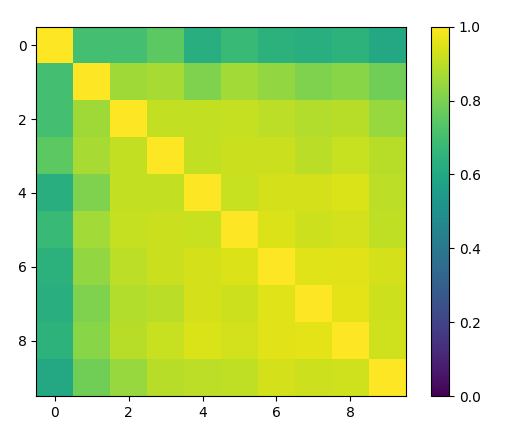}
    \subcaption{\small SCAFFOLD $\alpha$}
    \label{fig:cka_dirichlet_scaffold}
  \end{subfigure}%
  \caption{Similarity of the aggregated models via CKA on Dirichlet CIFAR-10}
  \label{fig:cka}
\end{figure*}

\begin{figure*}[!ht]
\hspace*{1cm}
  \begin{subfigure}[b]{.3\linewidth}
    \centering
    \includegraphics[height=3.0cm, width=4.0cm]{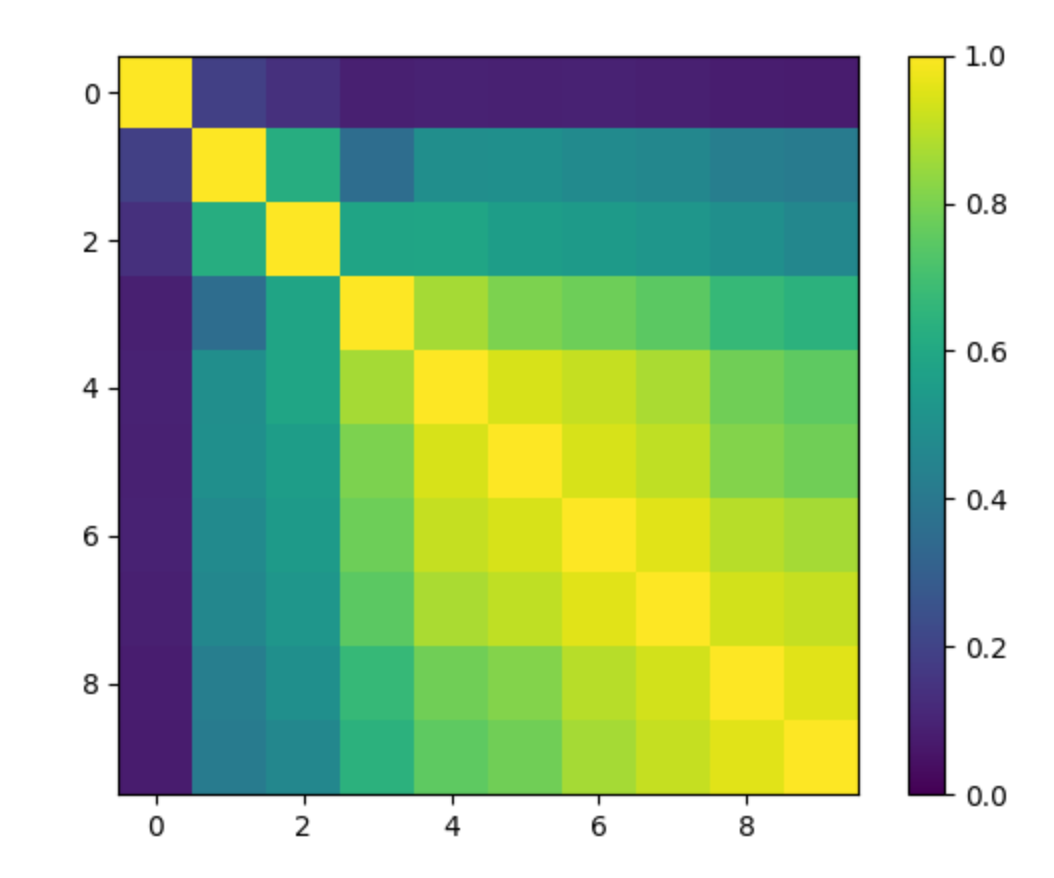}
    \subcaption{\small FedAVG $\beta$}
    \label{fig:cka_entropy_balance_fedavg}
  \end{subfigure}%
  \begin{subfigure}[b]{.3\linewidth}
    \centering
    \includegraphics[height=3.0cm, width=4.0cm]{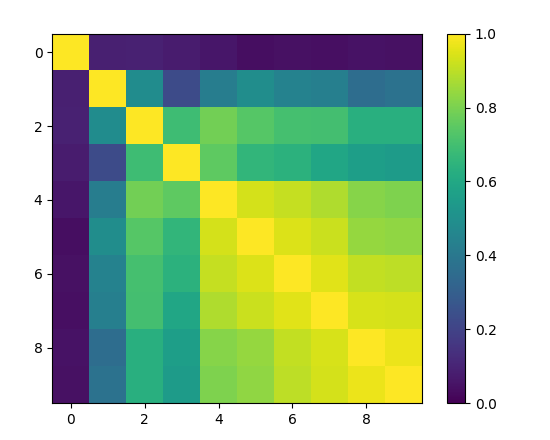}
    \subcaption{\small FedProx $\beta$}
    \label{fig:cka_entropy_balance_fedprox}
  \end{subfigure}%
  \begin{subfigure}[b]{.3\linewidth}
    \centering
    \includegraphics[height=2.9cm, width=4.0cm]{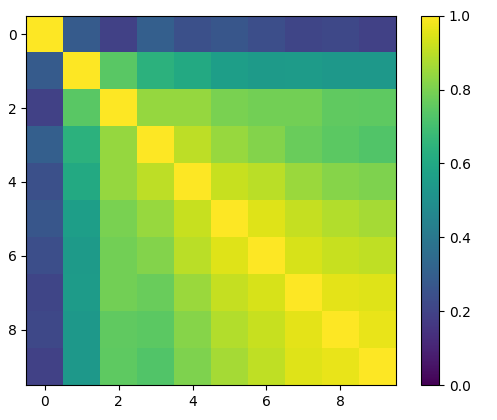}
    \subcaption{\small SCAFFOLD $\beta$}
    \label{fig:cka_entropy_balance_scaffold}
  \end{subfigure}
  \caption{Similarity of the aggregated models via CKA on FedSym, CIFAR-10}
  \label{fig:cka2}
\end{figure*}

\textbf{Comparison of the models' outputs via Centered Kernel Alignment.} In this experiment, we aim to measure the similarity between the generated models by various algorithms over the data distributions generated by the Dirichlet partitioning strategy and FedSym. By using the test dataset of CIFAR10 (10K images), we perform a cross-comparison of all ten aggregated (global) models for the values of $\alpha$ and $\beta$. Figures \ref{fig:cka_dirichlet_fedavg}, \ref{fig:cka_dirichlet_fedprox} and \ref{fig:cka_dirichlet_scaffold} suggest that the models generated by $\alpha \in [0.3, 1.0]$ produce very similar outputs, while figures \ref{fig:cka_entropy_balance_fedavg}, \ref{fig:cka_entropy_balance_fedprox} and \ref{fig:cka_entropy_balance_scaffold} show that the models trained on FedSym distributions differ gradually and more significantly from each other. We conclude that FedSym provides data distributions that incrementally challenge the FL algorithms and are a base for stronger algorithm comparisons.

\section{Constraints and opportunities for future work.}
\label{limitations}

The most important constraint of the method is that from a given dataset $D$ and $k$ learning agents, and $S$ samples per learning agent (client), it could utilize $k\times S$ training samples from $D$. Therefore, if $k\times S$ does not divide $|D|$, some of the training samples from $D$ may not be present in the final distributed training datasets. Another important constraint of the current algorithm is that the number of training samples should be equal among all classes e.g. $l$ should divide $k\times S$. In the future, one may examine the option to avoid both of these limitations and propose a method with an unequal number of samples per client and/or per class.

\section{Conclusion}
\label{conclusion}
\par The paper explores partitioning strategies for generating training datasets for decentralized machine learning to simulate non-IID (heterogeneous) data. The authors study the entropy balance of the generated datasets, as well as the performance of popular FL algorithms on the data partitions, to measure the difficulty of the datasets and the similarity of the trained models.
The proposed method - FedSym, guarantees the degree of data diversity and proves to challenge the algorithms for federated learning gradually. In the future, the work can be continued in at least two directions: first, perform a complete benchmark for the current state-of-the-art algorithms, and second, the proposed method can be further elaborated to provide non-symmetrical distributions while still guaranteeing the value of mean entropy balance for the generated datasets.

\section{Acknowledgment}
\selectlanguage{russian}
The result presented in this paper is part of the GATE project. The project has received funding from the EU’s Horizon 2020 Widespread-2018-2020 Teaming Phase 2 Programme under Grant Agreement No. 857155 and Operational Programme Science and Education for Smart Growth under Grant Agreement No. BG05M2OP001-1.003-0002-C01. The work was also supported by the CHIST-ERA grant CHIST-ERA-19-XAI-010, by MUR, FWF (grant No. I 5205), EPSRC (grant No. EP/V055712/1), NCN (grant No. 2020/02/Y/ST6/00064), ETAg (grant No. SLTAT21096), BNSF (grant No. KP-06-ДОО2/5).

\selectlanguage{english}
\small
\bibliographystyle{unsrt}
\bibliography{ref}

\begin{thebibliography}{10}

\bibitem{voigt2017eu}
Paul Voigt and Axel Von~dem Bussche.
\newblock The eu general data protection regulation (gdpr).
\newblock {\em A Practical Guide, 1st Ed., Cham: Springer International
  Publishing}, 10(3152676):10--5555, 2017.

\bibitem{DBLP:journals/corr/McMahanMRA16}
H.~Brendan McMahan, Eider Moore, Daniel Ramage, and Blaise~Ag{\"{u}}era
  y~Arcas.
\newblock Federated learning of deep networks using model averaging.
\newblock {\em CoRR}, abs/1602.05629, 2016.

\bibitem{li2018federated}
Tian Li, Anit~Kumar Sahu, Manzil Zaheer, Maziar Sanjabi, Ameet Talwalkar, and
  Virginia Smith.
\newblock Federated optimization in heterogeneous networks.
\newblock {\em arXiv preprint arXiv:1812.06127}, 2018.

\bibitem{karimireddy2020scaffold}
Sai~Praneeth Karimireddy, Satyen Kale, Mehryar Mohri, Sashank Reddi, Sebastian
  Stich, and Ananda~Theertha Suresh.
\newblock Scaffold: Stochastic controlled averaging for federated learning.
\newblock In {\em International Conference on Machine Learning}, pages
  5132--5143. PMLR, 2020.

\bibitem{pillutla2019robust}
Krishna Pillutla, Sham~M Kakade, and Zaid Harchaoui.
\newblock Robust aggregation for federated learning.
\newblock {\em arXiv preprint arXiv:1912.13445}, 2019.

\bibitem{reddi2020adaptive}
Sashank Reddi, Zachary Charles, Manzil Zaheer, Zachary Garrett, Keith Rush,
  Jakub Kone{\v{c}}n{\`y}, Sanjiv Kumar, and H~Brendan McMahan.
\newblock Adaptive federated optimization.
\newblock {\em arXiv preprint arXiv:2003.00295}, 2020.

\bibitem{xie2020multi}
Ming Xie, Guodong Long, Tao Shen, Tianyi Zhou, Xianzhi Wang, and Jing Jiang.
\newblock Multi-center federated learning.
\newblock {\em arXiv preprint arXiv:2005.01026}, 2020.

\bibitem{wang2020fedma}
Hongyi Wang, Mikhail Yurochkin, Yuekai Sun, Dimitris Papailiopoulos, and
  Yasaman Khazaeni.
\newblock Federated learning with matched averaging.
\newblock {\em arXiv preprint arXiv:2002.06440}, 2020.

\bibitem{lin2020ensemble}
Tao Lin, Lingjing Kong, Sebastian~U Stich, and Martin Jaggi.
\newblock Ensemble distillation for robust model fusion in federated learning.
\newblock {\em Advances in Neural Information Processing Systems},
  33:2351--2363, 2020.

\bibitem{mothukuri2021survey}
Viraaji Mothukuri, Reza~M Parizi, Seyedamin Pouriyeh, Yan Huang, Ali
  Dehghantanha, and Gautam Srivastava.
\newblock A survey on security and privacy of federated learning.
\newblock {\em Future Generation Computer Systems}, 115:619--640, 2021.

\bibitem{xu2021federated}
Jie Xu, Benjamin~S Glicksberg, Chang Su, Peter Walker, Jiang Bian, and Fei
  Wang.
\newblock Federated learning for healthcare informatics.
\newblock {\em Journal of Healthcare Informatics Research}, 5:1--19, 2021.

\bibitem{konevcny2016federated}
Jakub Kone{\v{c}}n{\`y}, H~Brendan McMahan, Felix~X Yu, Peter Richt{\'a}rik,
  Ananda~Theertha Suresh, and Dave Bacon.
\newblock Federated learning: Strategies for improving communication
  efficiency.
\newblock {\em arXiv preprint arXiv:1610.05492}, 2016.

\bibitem{huang2021personalized}
Yutao Huang, Lingyang Chu, Zirui Zhou, Lanjun Wang, Jiangchuan Liu, Jian Pei,
  and Yong Zhang.
\newblock Personalized cross-silo federated learning on non-iid data.
\newblock In {\em AAAI}, pages 7865--7873, 2021.

\bibitem{zhu2021federated}
Hangyu Zhu, Jinjin Xu, Shiqing Liu, and Yaochu Jin.
\newblock Federated learning on non-iid data: A survey.
\newblock {\em Neurocomputing}, 465:371--390, 2021.

\bibitem{DBLP:journals/corr/abs-2102-02079}
Qinbin Li, Yiqun Diao, Quan Chen, and Bingsheng He.
\newblock Federated learning on non-iid data silos: An experimental study.
\newblock {\em CoRR}, abs/2102.02079, 2021.

\bibitem{no_fear}
Mi~Luo, Fei Chen, Dapeng Hu, Yifan Zhang, Jian Liang, and Jiashi Feng.
\newblock No fear of heterogeneity: Classifier calibration for federated
  learning with non-iid data.
\newblock {\em CoRR}, abs/2106.05001, 2021.

\bibitem{wang2022unreasonable}
Jianyu Wang, Rudrajit Das, Gauri Joshi, Satyen Kale, Zheng Xu, and Tong Zhang.
\newblock On the unreasonable effectiveness of federated averaging with
  heterogeneous data.
\newblock {\em arXiv preprint arXiv:2206.04723}, 2022.

\bibitem{vahidian2022rethinking}
Saeed Vahidian, Mahdi Morafah, Chen Chen, Mubarak Shah, and Bill Lin.
\newblock Rethinking data heterogeneity in federated learning: Introducing a
  new notion and standard benchmarks.
\newblock In {\em Workshop on Federated Learning: Recent Advances and New
  Challenges (in Conjunction with NeurIPS 2022)}.

\bibitem{vajapeyam2014understanding}
Sriram Vajapeyam.
\newblock Understanding shannon's entropy metric for information.
\newblock {\em arXiv preprint arXiv:1405.2061}, 2014.

\bibitem{magurran2003measuring}
Anne~E Magurran.
\newblock {\em Measuring Biological Diversity}.
\newblock John Wiley \& Sons, 2003.

\bibitem{golden2020statistical}
Richard~M Golden.
\newblock {\em Statistical machine learning: A unified framework}.
\newblock Chapman and Hall/CRC, 2020.

\bibitem{McMahan2016}
Brendan McMahan, Eider Moore, Daniel Ramage, Seth Hampson, and Blaise~Aguera
  y~Arcas.
\newblock {Communication-Efficient Learning of Deep Networks from Decentralized
  Data}.
\newblock In {\em AISTATS}, pages 1273--1282, 2017.

\bibitem{zhao2018federated}
Yue Zhao, Meng Li, Liangzhen Lai, Naveen Suda, Damon Civin, and Vikas Chandra.
\newblock Federated learning with non-iid data.
\newblock {\em arXiv preprint arXiv:1806.00582}, 2018.

\bibitem{kairouz2019advances}
Peter Kairouz, H~Brendan McMahan, Brendan Avent, Aur{\'e}lien Bellet, Mehdi
  Bennis, Arjun~Nitin Bhagoji, Keith Bonawitz, Zachary Charles, Graham Cormode,
  Rachel Cummings, et~al.
\newblock Advances and open problems in federated learning.
\newblock {\em arXiv preprint arXiv:1912.04977}, 2019.

\bibitem{li2020lotteryfl}
Ang Li, Jingwei Sun, Binghui Wang, Lin Duan, Sicheng Li, Yiran Chen, and Hai
  Li.
\newblock Lotteryfl: Personalized and communication-efficient federated
  learning with lottery ticket hypothesis on non-iid datasets.
\newblock {\em arXiv preprint arXiv:2008.03371}, 2020.

\bibitem{qu2022rethinking}
Liangqiong Qu, Yuyin Zhou, Paul~Pu Liang, Yingda Xia, Feifei Wang, Ehsan Adeli,
  Li~Fei-Fei, and Daniel Rubin.
\newblock Rethinking architecture design for tackling data heterogeneity in
  federated learning.
\newblock In {\em Proceedings of the IEEE/CVF Conference on Computer Vision and
  Pattern Recognition}, pages 10061--10071, 2022.

\bibitem{zhang2022federated}
Jie Zhang, Zhiqi Li, Bo~Li, Jianghe Xu, Shuang Wu, Shouhong Ding, and Chao Wu.
\newblock Federated learning with label distribution skew via logits
  calibration.
\newblock In {\em International Conference on Machine Learning}, pages
  26311--26329. PMLR, 2022.

\bibitem{geyer2017differentially}
Robin~C Geyer, Tassilo Klein, and Moin Nabi.
\newblock Differentially private federated learning: A client level
  perspective.
\newblock {\em arXiv preprint arXiv:1712.07557}, 2017.

\bibitem{yurochkin2019bayesian}
Mikhail Yurochkin, Mayank Agarwal, Soumya Ghosh, Kristjan Greenewald, Nghia
  Hoang, and Yasaman Khazaeni.
\newblock Bayesian nonparametric federated learning of neural networks.
\newblock In {\em International Conference on Machine Learning}, pages
  7252--7261. PMLR, 2019.

\bibitem{wang2020federated}
Hongyi Wang, Mikhail Yurochkin, Yuekai Sun, Dimitris Papailiopoulos, and
  Yasaman Khazaeni.
\newblock Federated learning with matched averaging.
\newblock {\em arXiv preprint arXiv:2002.06440}, 2020.

\bibitem{hsu2019measuring}
Tzu-Ming~Harry Hsu, Hang Qi, and Matthew Brown.
\newblock Measuring the effects of non-identical data distribution for
  federated visual classification.
\newblock {\em arXiv preprint arXiv:1909.06335}, 2019.

\bibitem{li2020practical}
Qinbin Li, Bingsheng He, and Dawn Song.
\newblock Practical one-shot federated learning for cross-silo setting.
\newblock {\em arXiv preprint arXiv:2010.01017}, 2020.

\bibitem{wang2020tackling}
Jianyu Wang, Qinghua Liu, Hao Liang, Gauri Joshi, and H~Vincent Poor.
\newblock Tackling the objective inconsistency problem in heterogeneous
  federated optimization.
\newblock {\em Advances in neural information processing systems},
  33:7611--7623, 2020.

\bibitem{acar2021federated}
Durmus Alp~Emre Acar, Yue Zhao, Ramon~Matas Navarro, Matthew Mattina, Paul~N
  Whatmough, and Venkatesh Saligrama.
\newblock Federated learning based on dynamic regularization.
\newblock {\em arXiv preprint arXiv:2111.04263}, 2021.

\bibitem{awan2021contra}
Sana Awan, Bo~Luo, and Fengjun Li.
\newblock Contra: Defending against poisoning attacks in federated learning.
\newblock In {\em Computer Security--ESORICS 2021: 26th European Symposium on
  Research in Computer Security, Darmstadt, Germany, October 4--8, 2021,
  Proceedings, Part I 26}, pages 455--475. Springer, 2021.

\bibitem{fraboni2021clustered}
Yann Fraboni, Richard Vidal, Laetitia Kameni, and Marco Lorenzi.
\newblock Clustered sampling: Low-variance and improved representativity for
  clients selection in federated learning.
\newblock In {\em International Conference on Machine Learning}, pages
  3407--3416. PMLR, 2021.

\bibitem{gao2022feddc}
Liang Gao, Huazhu Fu, Li~Li, Yingwen Chen, Ming Xu, and Cheng-Zhong Xu.
\newblock Feddc: Federated learning with non-iid data via local drift
  decoupling and correction.
\newblock In {\em Proceedings of the IEEE/CVF Conference on Computer Vision and
  Pattern Recognition}, pages 10112--10121, 2022.

\bibitem{liu2022energy}
Xiaolan Liu, Yansha Deng, and Toktam Mahmoodi.
\newblock Energy efficient user scheduling for hybrid split and federated
  learning in wireless uav networks.
\newblock In {\em ICC 2022-IEEE International Conference on Communications},
  pages 1--6. IEEE, 2022.

\bibitem{sturluson2021fedrad}
Stef{\'a}n~P{\'a}ll Sturluson, Samuel Trew, Luis Mu{\~n}oz-Gonz{\'a}lez, Matei
  Grama, Jonathan Passerat-Palmbach, Daniel Rueckert, and Amir Alansary.
\newblock Fedrad: Federated robust adaptive distillation.
\newblock {\em arXiv preprint arXiv:2112.01405}, 2021.

\bibitem{canonne2020discrete}
Cl{\'e}ment~L Canonne, Gautam Kamath, and Thomas Steinke.
\newblock The discrete gaussian for differential privacy.
\newblock {\em Advances in Neural Information Processing Systems},
  33:15676--15688, 2020.

\bibitem{andrich2021rasch}
David Andrich.
\newblock The rasch distribution: A discrete, general form of the gauss
  distribution of uncertainty in scientific measurement.
\newblock {\em Measurement}, 173:108672, 2021.

\bibitem{stewart2009probability}
William~J Stewart.
\newblock {\em Probability, Markov chains, queues, and simulation: the
  mathematical basis of performance modeling}.
\newblock Princeton university press, 2009.

\bibitem{conrad2004probability}
Keith Conrad.
\newblock Probability distributions and maximum entropy.
\newblock {\em Entropy}, 6(452):10, 2004.

\end{thebibliography}


\begin{table*}[!ht]
\begin{center}
\hspace*{-1cm}
\begin{tabular}{|c|ccc|ccc|c|}
\hline
\multirow{2}{*}{H. Index} &
  \multicolumn{3}{c|}{FedSym} &
  \multicolumn{3}{c|}{Dirichlet} &
  \multicolumn{1}{l|}{SC\_NIID} \\ \cline{2-8} 
 &
  \multicolumn{1}{l|}{MNIST} &
  \multicolumn{1}{l|}{CIFAR10} &
  CINIC10 &
  \multicolumn{1}{l|}{MNIST} &
  \multicolumn{1}{l|}{CIFAR10} &
  CINIC10 &
  \multicolumn{1}{l|}{CIFAR10} \\ \hline
0.1 &
  \multicolumn{1}{l|}{$0.10 \pm 0.00$} &
  \multicolumn{1}{l|}{$0.10 \pm 0.00$} &
  $0.10 \pm 0.00$ &
  \multicolumn{1}{l|}{$0.37 \pm 0.15$} &
  \multicolumn{1}{l|}{$0.33 \pm 0.15$} &
  $0.33 \pm 0.18$ &
  $0.13 \pm 0.10$ \\ \hline
0.2 &
  \multicolumn{1}{l|}{$0.20 \pm 0.00$} &
  \multicolumn{1}{l|}{$0.20 \pm 0.00$} &
  $0.20 \pm 0.00$ &
  \multicolumn{1}{l|}{$0.52 \pm 0.09$} &
  \multicolumn{1}{l|}{$0.45 \pm 0.11$} &
  $0.38 \pm 0.14$ &
  $0.26 \pm 0.18 $ \\ \hline
0.3 &
  \multicolumn{1}{l|}{$0.30 \pm 0.00$} &
  \multicolumn{1}{l|}{$0.30 \pm 0.00$} &
  $0.30 \pm 0.00$ &
  \multicolumn{1}{l|}{$0.59 \pm 0.13$} &
  \multicolumn{1}{l|}{$0.48 \pm 0.08$} &
  $0.51 \pm 0.13$ &
  $0.36 \pm 0.09 $ \\ \hline
0.4 &
  \multicolumn{1}{l|}{$0.40 \pm 0.00$} &
  \multicolumn{1}{l|}{$0.40 \pm 0.00$} &
  $0.40 \pm 0.00$ &
  \multicolumn{1}{l|}{$0.67 \pm 0.07$} &
  \multicolumn{1}{l|}{$0.58 \pm 0.14$} &
  $0.59 \pm 0.10$ &
  $0.42 \pm 0.17 $ \\ \hline
0.5 &
  \multicolumn{1}{l|}{$0.50 \pm 0.00$} &
  \multicolumn{1}{l|}{$0.50 \pm 0.00$} &
  $0.50 \pm 0.00$ &
  \multicolumn{1}{l|}{$0.69 \pm 0.09$} &
  \multicolumn{1}{l|}{$0.63 \pm 0.11$} &
  $0.65 \pm 10$ &
  $0.39 \pm 0.15 $ \\ \hline
0.6 &
  \multicolumn{1}{l|}{$0.60 \pm 0.00$} &
  \multicolumn{1}{l|}{$0.60 \pm 0.00$} &
  $0.60 \pm 0.00$ &
  \multicolumn{1}{l|}{$0.76 \pm 0.09$} &
  \multicolumn{1}{l|}{$0.68 \pm 0.09$} &
  $0.70 \pm 0.06$ &
  $0.50 \pm 0.16 $ \\ \hline
0.7 &
  \multicolumn{1}{l|}{$0.70 \pm 0.00$} &
  \multicolumn{1}{l|}{$0.70 \pm 0.00$} &
  $0.70 \pm 0.00$ &
  \multicolumn{1}{l|}{$0.79 \pm 0.07$} &
  \multicolumn{1}{l|}{$0.72 \pm 0.10$} &
  $0.74 \pm 0.08$ &
  $0.45 \pm 0.11 $ \\ \hline
0.8 &
  \multicolumn{1}{l|}{$0.80 \pm 0.00$} &
  \multicolumn{1}{l|}{$0.80 \pm 0.00$} &
  $0.80 \pm 0.00$ &
  \multicolumn{1}{l|}{$0.81 \pm 0.08$} &
  \multicolumn{1}{l|}{$0.74 \pm 0.14$} &
  $0.78 \pm 0.08$ &
  $0.55 \pm 0.09$ \\ \hline
0.9 &
  \multicolumn{1}{l|}{$0.90 \pm 0.00$} &
  \multicolumn{1}{l|}{$0.90 \pm 0.00$} &
  $0.90 \pm 0.00$ &
  \multicolumn{1}{l|}{$0.82 \pm 0.07$} &
  \multicolumn{1}{l|}{$0.77 \pm 0.11$} &
  $0.77 \pm 0.07$ &
  $0.49 \pm 0.09 $ \\ \hline
1.0 &
  \multicolumn{1}{l|}{$1.00 \pm 0.00$} &
  \multicolumn{1}{l|}{$1.00 \pm 0.00$} &
  $1.00 \pm 0.00$ &
  \multicolumn{1}{l|}{$0.79 \pm 0.08$} &
  \multicolumn{1}{l|}{$0.85 \pm 0.05$} &
  $0.76 \pm 0.11$ &
  $0.53 \pm 0.19 $ \\ \hline
\end{tabular}
\end{center}
\caption{\smaller Mean and standard deviation of the calculated entropy balance for Dirichlet, SC\_NIID and FedSym methods for 10 clients, and for $\alpha, \beta \in [0.1, 1.0]$}
\label{table_eb}
\end{table*}

\section*{Appendix}
The Appendix is organized as follows: section \ref{section_eb} provides a study on the entropy balance of various data partitioning methods over popular machine learning datasets, section \ref{section_impact} demonstrates the relation between the entropy balance of these data distributions and performance of the algorithms for federated learning, and section \ref{section_cka} uses the CKA metric to showcase that the models trained on Dirichlet data partitions have fairly identical output, in contrast to the ones trained on FedSym data distributions. A final wrap-up is available in section \ref{section_conclusion}.

\section{The Entropy Balance as Dataset Difficulty Indicator}
\label{section_eb}

The primary objective of this work is to enhance the experimental component of the main paper, focusing on a more comprehensive study of data distributions derived from widely recognized machine learning datasets. Specifically, we utilize well-established datasets such as MNIST, CIFAR-10, and CINIC-10 to generate data distributions.

For each of these datasets, we generate data distributions for a collection of 10 learning agents. This is accomplished using diverse data partitioning techniques, including FedSym, Dirichlet, and the SC-NIID method introduced by \cite{qu2022rethinking}. Following this, we compute the mean entropy balance for all 10 data partitions and present our findings in table \ref{table_eb}.

It is important to note that the selection of 10 learning agents aligns with the standard benchmarking configuration in Federated Learning (FL) research\cite{no_fear}, a practice consistently employed by various researchers in the field. This benchmarking configuration ensures the comparability of our findings with other studies in this domain.

 \begin{figure}[!ht]
    \hspace{-0.5cm}
    \centering
    \includegraphics[height=5.5cm, width=9.5cm]{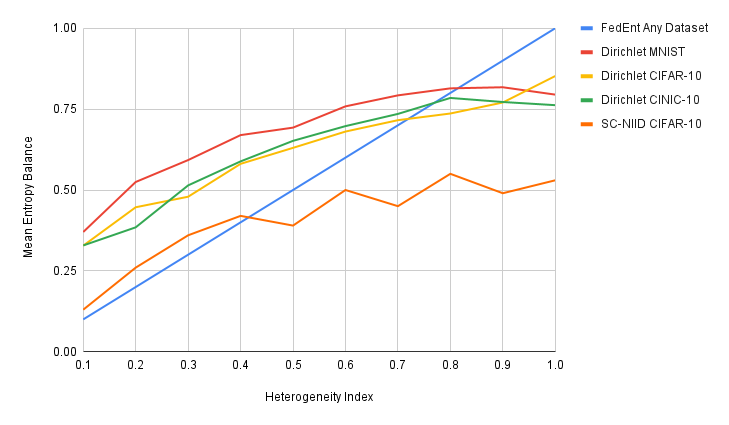}
    \caption{Mean Entropy Balance per Heterogeneity Index}
    \label{fig:mean_ent}
\end{figure}

 Figure \ref{fig:mean_ent} depicts the correlation between the heterogeneity index, represented on the X-axis, and the derived mean entropy balance indicated on the Y-axis. It is evident that the FedSym method ensures the entropy balance of the resultant data partitions. Contrastingly, there seems to be an absence of a direct correlation between the value of the heterogeneity index and the final entropy of the datasets when using alternative methods.

More specifically, in the case of the Dirichlet method for data partitioning, the heterogeneity index $\alpha \in [0.6, 1.0]$ yields data partitions possessing identical or nearly identical entropy. The comprehensive range of the entropy balances acquired via the Dirichlet method resides in the interval of approximately $(0.3, 0.8)$. Meanwhile, for the SC-NIID method, a derivative of the Dirichlet method, the entropy balance lies within the approximate bounds of $(0.1, 0.55)$.

Based on these findings, we infer that only the FedSym method delivers a broad spectrum of distributions, each with a guaranteed degree of heterogeneity. This observation underscores the value of the FedSym method in providing reliable entropy balances across varying degrees of dataset heterogeneity.

\begin{figure*}[!hb]
\hspace*{-1.7cm}
  \begin{subfigure}[b]{.4\linewidth}
    \centering
    \includegraphics[height=4cm, width=4.5cm]{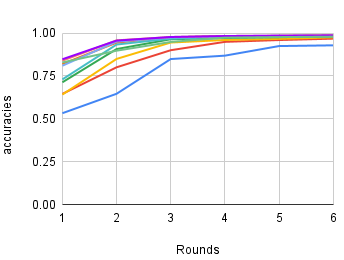}
    \subcaption{FedAVG on Dirichlet distributions}
    \label{fig:avg_dir_acc_mnist}
  \end{subfigure}%
  \begin{subfigure}[b]{.4\linewidth}
    \centering
    \includegraphics[height=4cm, width=4.5cm]{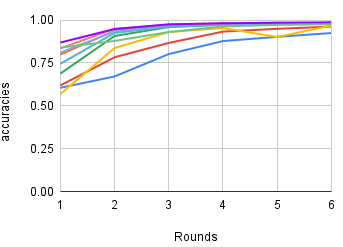}
    \subcaption{FedProx on Dirichlet distributions}
    \label{fig:prox_dir_acc_mnist}
  \end{subfigure}%
  \begin{subfigure}[b]{.4\linewidth}
    \centering
    \includegraphics[height=4cm, width=5.0cm]{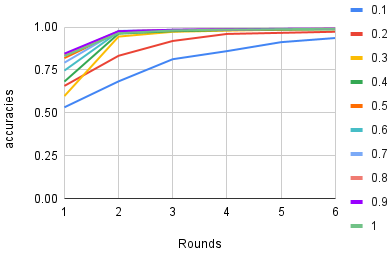}
    \subcaption{SCAFFOLD on Dirichlet distributions}
    \label{fig:scaffold_dir_acc_mnist}
  \end{subfigure}
\end{figure*}

\begin{figure*}[!hb]
\hspace*{-1.7cm}
  \begin{subfigure}[b]{.4\linewidth}
    \centering
    \includegraphics[height=4cm, width=4.5cm]{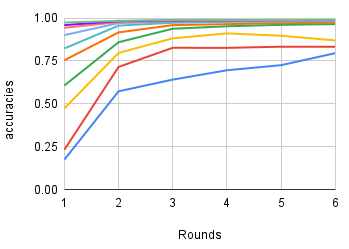}
    \subcaption{FedAVG on FedSym distributions}
    \label{fig:avg_ent_acc_mnist}
  \end{subfigure}%
  \begin{subfigure}[b]{.4\linewidth}
    \centering
    \includegraphics[height=4.0cm, width=4.5cm]{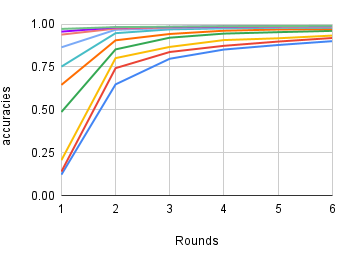}
    \subcaption{FedProx on FedSym distributions}
    \label{fig:prox_ent_acc_mnist}
  \end{subfigure}%
  \begin{subfigure}[b]{.4\linewidth}
    \centering
    \includegraphics[height=4cm, width=5.0cm]{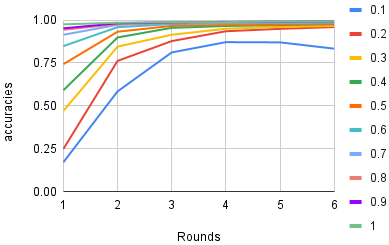}
    \subcaption{SCAFFOLD on FedSym distributions}
    \label{fig:scaffold_ent_acc_mnist}
  \end{subfigure}
\end{figure*}

\textbf{Experimental setup and computational requirements}: To replicate the outcomes of this study, the utilization of a Graphics Processing Unit (GPU) is not required. All of the data partitioning strategies implemented, including FedSym, can be efficiently conducted using conventional CPUs. Hence, the necessary computational power for executing these methods is readily available in most standard computing environments.

\section{Impact of Entropy Balance on Federated Learning Algorithms' Performance}
\label{section_impact}

This section aims to describe further the second experiment from our main paper, provide additional technical details for the benchmark execution and expand the results on another popular dataset, namely the MNIST. The advantage of using MNIST is that the experiment can be reproduced more easily on low-end GPUs. Although the model used in this experiment is more than 200 times smaller (45K vs. 10M trainable parameters), we discover identical results - the distributions created by FedSym are gradually challenging the FL algorithms and provide an environment with controlled difficulty.

\textbf{Experimental setup and computational requirements.} The present experiment includes a set of 10 data distributions for 10 learning agents. Each data distribution corresponds to a value of the heterogeneity index $\alpha \in [0.1, 1.0]$. As mentioned above, the used data model has a CNN feature extractor and 3-level fully-connected classifier, with a total number of 45K trainable parameters. We perform training in 6 communication rounds $\times$ 10 local epochs for each round and each learner. The test accuracy of the aggregated (global) model after each round is stored and shown in the charts of figures \ref{fig:avg_dir_acc_mnist} to \ref{fig:scaffold_dir_acc_mnist}, using the FL algorithms FedAvg, FedProx, and SCAFFOLD. The identical experiment is also performed for distributions generated by the FedSym method, and the corresponding results are visualized in figures \ref{fig:avg_ent_acc_mnist} to \ref{fig:scaffold_ent_acc_mnist}.  As the architecture of the trained models is fairly minimal, the computational requirements are not significant and can be performed even on low-end graphics accelerators.

\textbf{Results interpretation.} In the previous experiment, we observed that the distributions generated by FedSym demonstrate a gradual increase in entropy values. This is in stark contrast with the datasets generated via the Dirichlet method, which exhibited restrained fluctuations in the mean entropy balance. As a result, we see that the accuracy of the global models trained on Dirichlet distributions has identical behavior (cf. figures \ref{fig:avg_dir_acc_mnist} to \ref{fig:scaffold_dir_acc_mnist}). On the other hand, the benchmarks of all three algorithms on FedSym distributions have distinguishable and gradually increasing performance that corresponds to the entropy of the data partitions, which is guaranteed by the heterogeneity index $\beta \in [0.1, 1.0]$.

\begin{figure*}[!ht]
  \begin{subfigure}[b]{.35\linewidth}
    \centering
    \includegraphics[height=4cm, width=5cm]{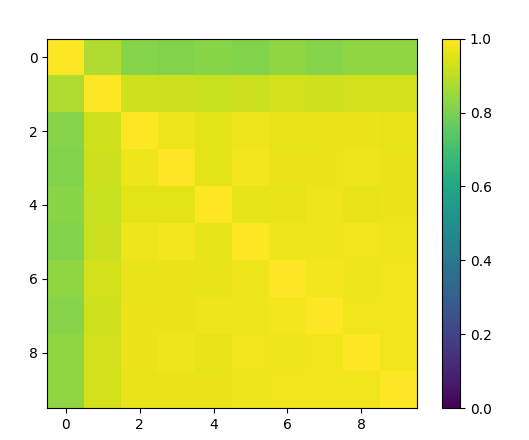}
    \subcaption{\small  On Dirichlet distributions}
    \label{fig:cka_dir_fedavg_mnist}
  \end{subfigure}%
  \begin{subfigure}[b]{.35\linewidth}
    \centering
    \includegraphics[height=4cm, width=5cm]{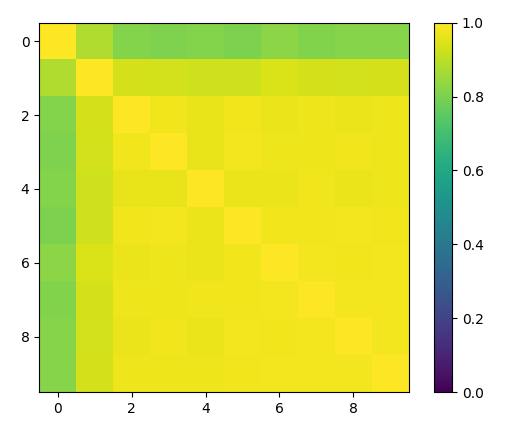}
    \subcaption{\small On Dirichlet distributions}
    \label{fig:cka_dir_fedprox_mnist}
  \end{subfigure}%
  \begin{subfigure}[b]{.35\linewidth}
    \centering
    \includegraphics[height=4cm, width=5cm]{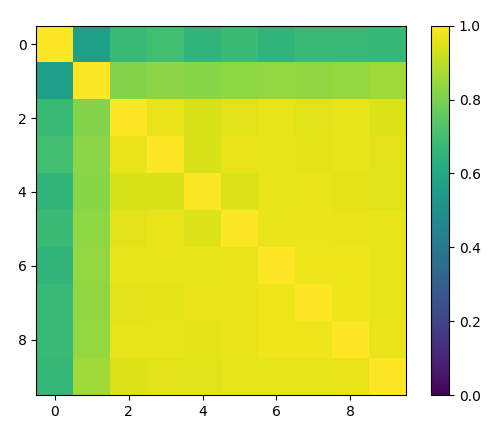}
    \subcaption{\small On Dirichlet distributions}
    \label{fig:cka_dir_scaffold_mnist}
  \end{subfigure}%
  \caption{Output similarity of the global models of FedAVG, FedProx, and Scaffold trained MNIST partitioend by drirchlet (CKA measure)}
  \label{fig:cka_dir_models_mnist}
\end{figure*}

\begin{figure*}[!ht]
  \begin{subfigure}[b]{.35\linewidth}
    \centering
    \includegraphics[height=4cm, width=5cm]{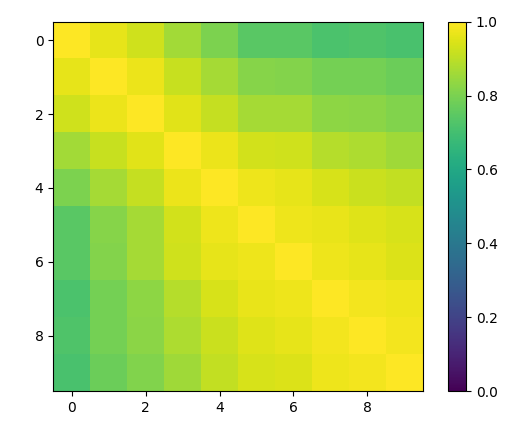}
    \subcaption{\small On FedSym distributions}
    \label{fig:cka_ent_fedavg_mnist}
  \end{subfigure}%
  \begin{subfigure}[b]{.35\linewidth}
    \centering
    \includegraphics[height=4cm, width=5cm]{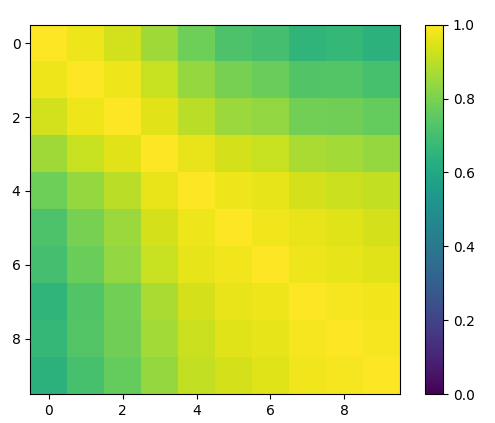}
    \subcaption{\small On FedSym distributions}
    \label{fig:cka_ent_fedprox_mnist}
  \end{subfigure}%
  \begin{subfigure}[b]{.35\linewidth}
    \centering
    \includegraphics[height=4cm, width=5cm]{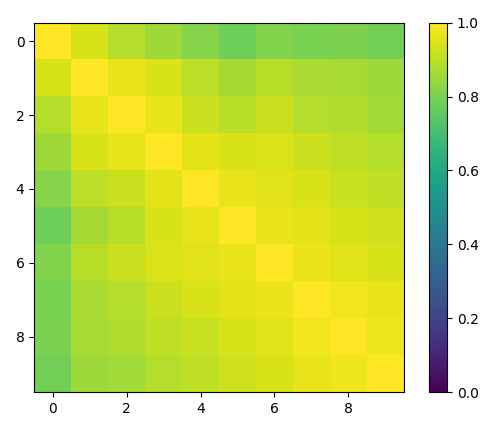}
    \subcaption{\small On FedSym distributions}
    \label{fig:cka_ent_scaffold_mnist}
  \end{subfigure}
  \caption{Output similarity of the global models of FedAVG, FedProx, and Scaffold trained on MNIST trained MNIST partitioned using FedSym (CKA measure)}
  \label{fig:cka_ent_models_mnist}
\end{figure*}

\section{Comparison of the models' outputs via Centered Kernel Alignment}
\label{section_cka}
In order to simplify the computational requirements of our experiments, we provide a study based on the simpler MNIST dataset using a light CNN network described in section \ref{section_impact}. The aim of this experiment is to provide a cross-comparison of all 10 global models, trained on distributions with different values of the heterogeneity indexes $\alpha, \beta \in [0.1, 1.0]$. We build a heatmap using the Centered Kernel Alignment (CKA) measure, where each column and row corresponds to one of the trained global models. The lighter colors mean more identical model output, while the darker ones indicate a difference in the model output. 

\textbf{Experimental setup and computational requirements}: The experiment involves a simple forward pass of all $10 \times 2 \times 3 = 60$ stored global models from the section \ref{section_impact} on the test set of the MNIST data. As each of the models possess only 45K trainable parameters, a graphic accelerator is not necessary, and the experiment can be executed on any decent computer system.

\textbf{Results interpretation.} As seen in figures \ref{fig:cka_dir_fedavg_mnist}, \ref{fig:cka_dir_fedprox_mnist}, and \ref{fig:cka_dir_scaffold_mnist} the models trained on Dirichlet distributions have fairly identical outputs, which indicates that the values of the heterogeneity index $\alpha$ have less importance to the final model. On the other hand, figures \ref{fig:cka_ent_fedavg_mnist}, \ref{fig:cka_ent_fedprox_mnist}, and \ref{fig:cka_ent_scaffold_mnist} are dominated by the darker values of the similarity metric, proving that the models trained on FedSym data partitions are distinguishable and each of the partitions, generated by the heterogeneity index $\beta \in [0.1, 1.0]$ produces a model with unique output.

\section{Conclusion}
\label{section_conclusion}
The additional set of experiments provided in this appendix aims to confirm the relation between entropy of datasets and the performance of federated machine learning. By studying the entropy of various data partitioning methods, we showcase that the performance of the algorithms for federated learning is directly tied to the values of the entropy balance. Furthermore, we demonstrate that the data distributions generated by the proposed method FedSym provide a consistent and gradual challenge to the algorithms for FL. And last but not least, this additional set of experiments is designed to be executable on graphic accelerators with modest computational capabilities.

\end{document}